%% file: main_arxiv.tex
\documentclass[11pt]{article}

\usepackage[preprint]{acl}

\usepackage{times}
\usepackage{latexsym}

\usepackage[T1]{fontenc}

\usepackage[utf8]{inputenc}

\usepackage{microtype}

\usepackage{inconsolata}

\usepackage{graphicx}
\usepackage{booktabs}
\usepackage{amssymb}
\usepackage{amsmath}
\usepackage{hyperref}
\usepackage{multirow}
\usepackage[most]{tcolorbox}

\title{The Cases LJP Never Sees: Prosecution Decision Prediction for More Complete Criminal Liability Assessment}

\author{
  \textbf{Junyu Lu\textsuperscript{1}},
  \textbf{Qi Wei\textsuperscript{2}},
  \textbf{Peishuo Zheng\textsuperscript{1}},
  \textbf{Jie Zhang\textsuperscript{3}},
\\
  \textbf{Hui Huang\textsuperscript{4},}
  \textbf{Qianru Wang\textsuperscript{5},}
  \textbf{Chuan Xiao\textsuperscript{2},}
  \textbf{Jianbin Qin\textsuperscript{1},}
  \textbf{Shuyuan Zheng\textsuperscript{2}\thanks{Correspondence to: Shuyuan Zheng <zheng@ist.osaka-u.ac.jp>.}}
\\
  \textsuperscript{1}Beijing Institute of Technology, Zhuhai,
  \textsuperscript{2}Osaka University
\\
  \textsuperscript{3}Xi'an University of Technology,
  \textsuperscript{4}Institute of Science Tokyo,
  \textsuperscript{5}City University of Hong Kong
}

\input{savespace}

\begin{document}
\maketitle

\input{sections/0_Abstract}
\input{sections/intro_zheng}
\input{sections/2_Related_Works}
\input{sections/3_PDP}
\input{sections/4_Experiments}
\input{sections/5_Conclusion}
\input{sections/Limitations}

\bibliography{reference}

\clearpage
\appendix
\input{sections/Appendix}

\end{document}

%% file: sections/0_Abstract.tex

\begin{abstract}
Legal Judgment Prediction (LJP) has become a core benchmark for evaluating AI in the criminal legal domain, but it only sees criminal cases that have already passed prosecutorial review and been formally indicted. 
As a result, LJP leaves a substantial blind spot in assessing criminal liability, overlooking cases involving insufficient evidence, no criminal liability, or guilt exempted from punishment.
To fill this gap, we propose \textbf{Prosecution Decision Prediction (PDP)}, the first Legal AI task built around prosecutorial review, which classifies each case into prosecution or one of three non-prosecution decisions and reflects legal AI's capabilities in evidence evaluation, legal subsumption, and value-based discretion. 
We further construct \textbf{PDP-Bench}, a benchmark of 4{,}630 real Chinese prosecutorial decisions spanning 190 charges. Extensive experiments show that state-of-the-art LLMs perform substantially worse on PDP than on LJP and that mainstream enhancement routes fail to close the gap. Moreover, controlled RLVR interventions show that simple outcome rewards fail to produce generalizable PDP discrimination.\footnote{PDP-Bench is available at \url{https://huggingface.co/datasets/Julian2002/PDP-Bench}, and the code is available at \url{https://github.com/lu-jun-yu/PDP}.}
\end{abstract}

%% file: sections/intro_zheng.tex
\section{Introduction}

Legal Judgment Prediction (LJP) is a representative task of Legal AI that predicts judicial outcomes directly from case facts~\citep{zhong2018overview,xiao2018cail}. In the criminal domain, it takes a case's fact description as input and predicts the applicable law articles, charges, and prison terms~\citep{xiao2018cail}. By automating this prediction, LJP gives users a preview of likely outcomes, improving judicial efficiency and transparency~\citep{cui2023survey}. It has thus become a core benchmark for evaluating the legal reasoning ability of AI systems~\citep{zhang2024beyond}.

\begin{figure*}[t]
    \centering
    \includegraphics[width=\textwidth]{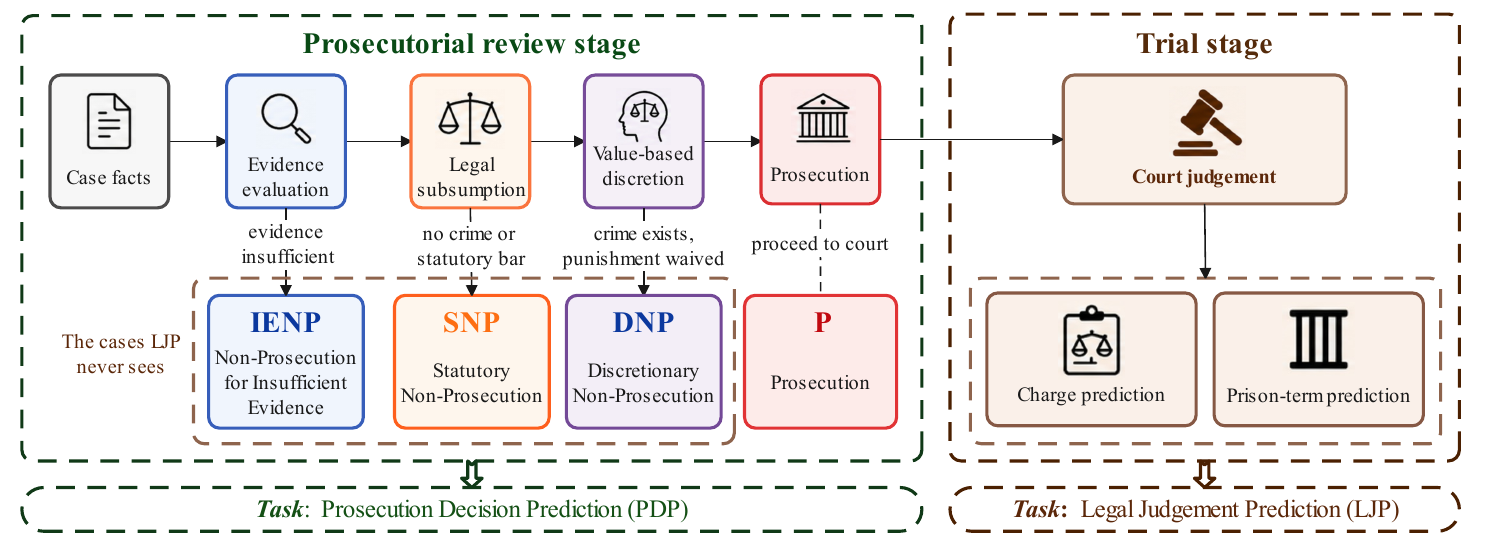}
    \caption{Prosecutorial-review-stage PDP reveals the cases trial-stage LJP never sees.}
    \label{fig:two-stage}
\end{figure*}

However, criminal LJP only sees cases that have already passed prosecutorial review and been formally indicted. 
As illustrated in Figure~\ref{fig:two-stage}, in typical criminal justice systems such as those of China and Japan, the procuratorate conducts prosecutorial review before trial to assess whether the suspect should bear criminal liability, and decides whether to file a prosecution. 
Cases found to involve insufficient evidence, no criminal liability, or guilt that exempts from punishment are diverted out of trial via non-prosecution decisions. 
Conversely, because prosecutorial review has already largely settled the question of criminal liability, the overwhelming majority of indicted cases are eventually convicted. 
For example, in China, the acquittal rate in public-prosecution cases has remained below 0.1\% since 2010 according to official judicial statistics and empirical studies~\citep{duihua2022acquittal,liang2023not}.
Existing LJP research therefore focuses on predicting specific charges and sentences, rather than whether the defendant should be held criminally liable at all~\citep{xiao2018cail,zhong2018overview,feng2022legal}.

As a result, LJP leaves a substantial blind spot in evaluating Legal AI's criminal-liability prediction. 
As Figure~\ref{fig:two-stage} shows, a complete criminal liability assessment should cover four outcomes: 
(1) \textbf{Guilty With Punishment}; 
(2) \textbf{Insufficient Evidence}, where the evidence does not meet the legal threshold for conviction; 
(3) \textbf{Not Guilty}, where the conduct is not criminal or liability is legally excluded; 
and (4) \textbf{Guilty Without Punishment}, where the conduct constitutes a crime but punishment is waived as a matter of discretion. 
LJP focuses on the first while overlooking the other three, with severe consequences: recent studies show that on artificially constructed innocent cases, state-of-the-art (SOTA) legal LLMs almost overwhelmingly return guilty verdicts~\citep{zhang2024beyond}.
Prosecutorial review, by contrast, naturally covers the three missing outcomes, offering a natural anchor for a more complete assessment of criminal liability; 
yet no Legal AI research has so far targeted this stage.

To fill this gap, we propose \textbf{Prosecution Decision Prediction (PDP)}, the first Legal AI task built around prosecutorial review. 
As shown in Figure~\ref{fig:two-stage}, PDP performs 4-way classification over the case facts, classifying each case into one of: \textbf{Non-Prosecution for Insufficient Evidence} (\textsc{IENP}), \textbf{Statutory Non-Prosecution} (\textsc{SNP}, implying Not Guilty), \textbf{Discretionary Non-Prosecution} (\textsc{DNP}, implying Guilty Without Punishment), and \textbf{Prosecution} (P, implying Guilty With Punishment). PDP thus restores the three criminal-liability outcomes LJP misses, and predicting them requires strong capabilities in \textbf{evidence evaluation}, \textbf{legal subsumption}, and \textbf{value-based discretion}---posing a comprehensive challenge to legal reasoning.

Building on this task, we construct \textbf{PDP-Bench}, the first PDP benchmark. 
We manually collect all prosecution and non-prosecution decisions released by procuratorates across China on their official websites up to March 2026, and curate 4,630 cases covering 190 criminal charges under legal-expert verification and refinement.

We then empirically analyze SOTA LLMs on PDP-Bench and drive the following findings. 
First, SOTA LLMs perform substantially worse on PDP than on LJP, highlighting the task's unique challenge. 
Second, three mainstream enhancement routes for LLMs fail to substantially improve PDP accuracy. 
Finally, since reinforcement learning with verifiable rewards (RLVR) has emerged as the dominant paradigm for boosting LLM reasoning~\citep{guo2025deepseekr1,lambert2024tulu3}, we show that simple outcome-reward RLVR fails to produce generalizable PDP discrimination, either amplifying target-class priors or providing too little signal to learn difficult legal boundaries.

Our contributions are threefold:
\begin{itemize}
    \item We propose PDP, a novel legal task complementing LJP by covering its three missing criminal-liability outcomes and evaluating legal AI's capabilities in evidence evaluation, legal subsumption, and value-based discretion.
    \item We construct a realistic PDP testbed, PDP-Bench, built from real-world prosecution and non-prosecution decisions of Chinese procuratorates.
    \item We provide extensive empirical analyses on PDP-Bench, demonstrating the substantial challenge PDP poses to current Legal AI.
\end{itemize}

%% file: sections/2_Related_Works.tex
\section{Related Work}


\paragraph{Legal AI and Legal Benchmarks.}
Legal AI research has explored diverse NLP-based tasks, including legal judgment prediction, legal entity recognition and classification, legal question answering, and summarization \citep{zhong2020nlp}. Prior work has also developed legal-domain models and benchmarks: LEGAL-BERT and Lawformer adapt pretrained models to legal text and long legal documents, while LexGLUE consolidates legal language understanding tasks and recent LLM benchmarks evaluate legal knowledge, reasoning, and agentic workflows \citep{chalkidis2020legalbert, xiao2021lawformer, chalkidis2022lexglue, guha2023legalbench, fei2024lawbench, li2024lexeval, li2025legalagentbench}.

\paragraph{Trial-Stage Legal Judgment Prediction.}
LJP typically predicts court-stage outcomes such as charges, applicable law articles, and penalties \citep{feng2022legaljudgment}. Early work studied judgment prediction in European and English court decisions \citep{aletras2016predicting, chalkidis2019neural}; then Chinese criminal LJP developed around charge prediction, legal-article prediction, and sentencing on court-judgment datasets such as CAIL2018 \citep{luo2017learning, xiao2018cail, zhong2018overview}. Subsequent studies address label sparsity and decision-boundary ambiguity through few-shot charge prediction, judgment-element dependency modeling, confusing-law discrimination, and contrastive LJP \citep{hu2018fewshot, zhong2018topjudge, xu2020distinguish, gan2023contrastive}; other studies improve rationale use, multi-defendant reasoning, LLM discrimination, and legal fact prediction \citep{ye2022rationale, yang2023multidefendant, huang2024cmdl, deng2024enabling, liu2025legalfact}. Despite methodological differences, these works share the same evaluation object: cases that have already reached trial. Even work adding acquittal labels expands the trial-stage label space rather than shifting the sampling source \citep{zhang2024beyond}.

\paragraph{Prosecutorial-Stage Tasks.}
One line uses Taiwanese prosecutorial materials for indictment-based knowledge management, error analysis, and prediction of processed outcomes or punishment attributes \citep{chien2024legalprosecutors, sun2023newhorizons, chien2024prosecutorial}. A second line studies prosecutorial decision support, including case screening, race masking in charging, rearrest-risk triage, and case prioritization under jurisdiction-specific governance settings \citep{chen2018algorithms,chohlaswood2021blind,sobrino2026designing}. A third line analyzes default prosecution tendencies in LLM-generated criminal-law memoranda \citep{pulvino2026hiding}. Existing procuratorate-related studies therefore cover knowledge management, governance support, and risk analysis, but not a multi-class NLP benchmark over Chinese prosecutorial-review documents.

%% file: sections/3_PDP.tex
\section{PDP}
\label{sec:pdp}

\subsection{Task Definition}

\textbf{Prosecution Decision Prediction (PDP)} is a four-way classification task for Chinese criminal cases at the prosecutorial-review stage. Given a case represented by three structured input fields---\textbf{Suspect Information}, \textbf{Procedural Information}, and \textbf{Factual Information} (defined in Table~\ref{tab:pdp_fields})---the model predicts $\hat{y}=f(x)\in\mathcal{Y}$ where $x=(x_s,x_p,x_f)$ and $\mathcal{Y}=\{\text{IENP},\text{SNP},\text{DNP},\text{P}\}$.

\subsection{PDP-Bench Construction}

\begin{figure}[t]
    \centering
    \includegraphics[width=\linewidth]{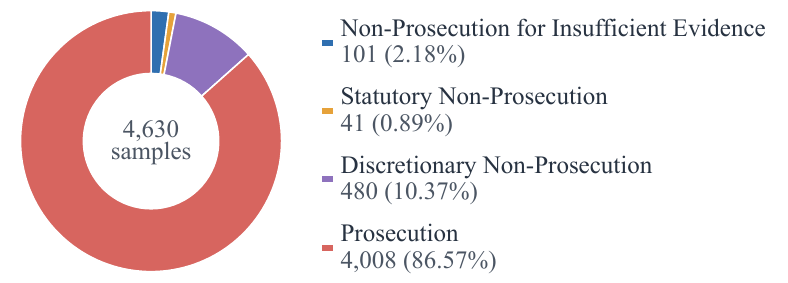}
    \caption{Overview of PDP-Bench.}
    \label{fig:pdp_overview}
\end{figure}

\begin{figure*}[t]
    \centering
    \includegraphics[width=\textwidth]{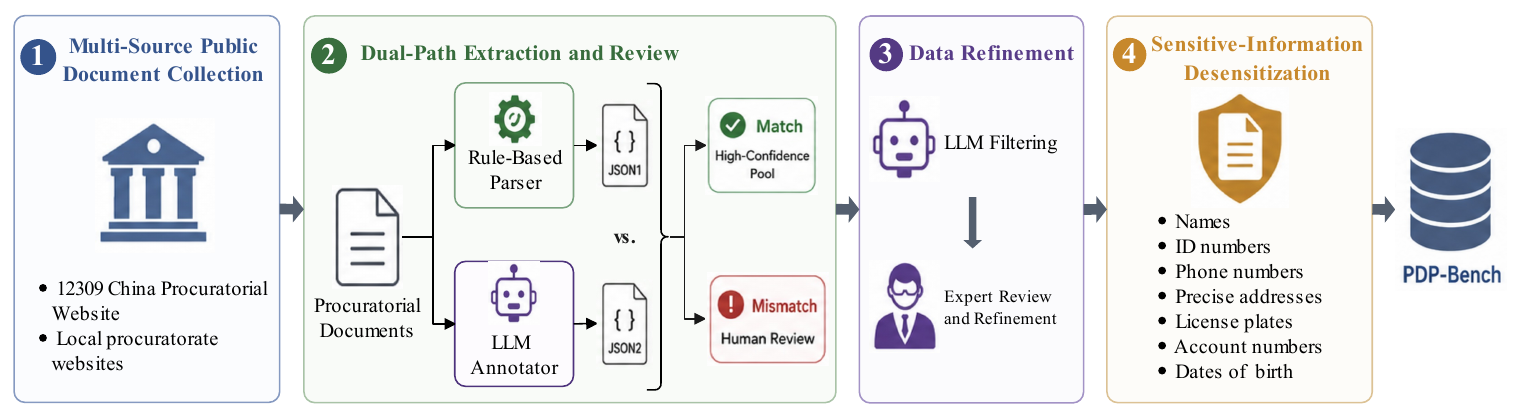}
    \caption{PDP-Bench dataset construction pipeline.}
    \label{fig:pdp_construction}
\end{figure*}

We construct \textbf{PDP-Bench}, a PDP benchmark curated entirely from publicly released Chinese prosecutorial documents so that it can be openly distributed and independently verified through the source URL retained in every sample. It contains 4{,}630 cases spanning January 2014 to March 2026 across all 31 provincial-level administrative regions of mainland China. Table~\ref{tab:pdp_fields} lists the per-sample fields and Figure~\ref{fig:pdp_overview} shows the label distribution. By design, we do not artificially rebalance the four classes: the resulting long-tail distribution reflects the empirical proportions observed in publicly available prosecutorial decisions, which we view as the correct evaluation regime---test data should reflect deployment-time class proportions rather than be artificially balanced. Figure~\ref{fig:pdp_construction} summarizes the four-stage construction pipeline.

\begin{table}[t]
\centering
\small
\begin{tabular}{@{}l p{0.58\columnwidth}@{}}
\toprule
\textbf{Field} & \textbf{Description} \\
\midrule
Case ID & Document case number \\
Metadata & Document date and province \\
Suspect Information & Suspect identity, coercive measures, and prior offenses \\
Procedural Information & Investigation, transfer for prosecutorial review, supplementary investigation, period extensions, and rights notification \\
Factual Information & Structured case facts and necessary evidence \\
Cited Law Articles & Applicable articles \\
Prosecution Decision & One of the four PDP labels \\
Original Reasoning & Original prosecutorial review opinion and decision \\
Source URL & Publicly accessible document URL \\
\bottomrule
\end{tabular}
\caption{Fields provided in each PDP-Bench sample.}
\label{tab:pdp_fields}
\end{table}

\paragraph{Stage 1: Multi-source public document collection.}
We collect indictments and non-prosecution decisions from 12309 China Procuratorial Network and local procuratorate websites. Because public documents are scattered across regions, platforms, and page formats, collection involves source verification, format normalization, body-text extraction, malformed-sample filtering, resumable organization, and cross-source deduplication.

\paragraph{Stage 2: Dual-track extraction and review.}
For the same document pool, we run two independent extraction tracks in parallel: (i) a regex- and rule-based parser that extracts the text fields under a predefined schema; and (ii) an LLM annotator that produces structured outputs under the same schema. We then perform full-set consistency checking. Samples whose two outputs agree enter a high-confidence pool, while discrepant samples are sent to human review. Reviewers compare the source document with both extracted versions, check field completeness, factual contamination by reasoning or conclusions, label correctness, and whether cited law articles are grounded in the source, and retain the higher-quality version. The corresponding extraction and review prompts are provided in Appendix~\ref{sec:app_data_prompts}.

\paragraph{Stage 3: Data refinement.}
We require every PDP-Bench sample to satisfy three data quality criteria, each targeting a specific failure mode observed in raw prosecutorial decisions:
 
\begin{itemize}
    \item \textbf{Fact-Evidence Formatting.} The evidentiary information of every case must be presented in a unified evidence-list format, since some prosecutorial decisions describe evidence in an unstructured form. For non-compliant cases, two legal experts reorganize the evidentiary information into the standardized evidence-list format.
    \item \textbf{Fact-Label Consistency.} The factual description must be substantively consistent with the conclusion of prosecutorial review, since a small fraction of raw documents contain overly brief facts sections that fail to support the corresponding decision. For non-compliant cases, two legal experts make minimally necessary additions while objectively preserving the original meaning of the facts section from a professional legal standpoint.
    \item \textbf{No Data Leakage.} The facts section must contain no extraneous information from which the prosecution decision can be inferred. For example, formulations such as ``the above evidence was lawfully collected, is objectively true, and is sufficient to establish the alleged facts'' allow one to rule out the insufficient-evidence non-prosecution class. For non-compliant cases, two legal experts remove such leakage-inducing content.
\end{itemize}
 
Because PDP-Bench contains 4{,}630 cases, exhaustive manual screening is impractical. We therefore use the Chinese SOTA LLM GLM-5.1 to automatically flag all cases violating any of the three criteria. The legal experts then perform secondary review on the flagged cases and refine those that remain non-compliant under human judgment.

\paragraph{Stage 4: Sensitive-information de-identification.}
We apply minimally necessary de-identification to all samples, using an LLM annotator with rule-based fallback. The procedure covers personally identifiable information in the suspect, procedural, factual, and original-reasoning fields, including names, national ID numbers, phone numbers, precise addresses, license plate numbers, account numbers, and dates of birth. The scope is strictly limited to identifying information. Legally relevant factual structure, monetary amounts, quantities, time expressions, charges, procedural events, law articles, and dispositions are preserved verbatim. The de-identification prompt is in Appendix~\ref{sec:app_data_prompts}.

%% file: sections/4_Experiments.tex
\begin{table*}[!t]
\centering
\small
\renewcommand{\arraystretch}{1.08}
\resizebox{\textwidth}{!}{%
\begin{tabular}{@{}c*{10}{c}@{}}
\toprule
\multirow{2}{*}{\textbf{Model}} & \multicolumn{1}{c}{\textbf{CAIL}} & \multicolumn{4}{c}{\textbf{PDP L1}} & \multicolumn{5}{c}{\textbf{PDP L2}} \\
\cmidrule(lr){2-2}\cmidrule(lr){3-6}\cmidrule(l){7-11}
& \textbf{Macro-F1} & \textbf{Macro-F1} & \textbf{Micro-F1} & \textbf{NP} & \textbf{P} & \textbf{Macro-F1} & \textbf{Micro-F1} & \textbf{IENP} & \textbf{SNP} & \textbf{DNP} \\
\midrule
GPT-5.4 & 0.6806 & 0.6899 & 0.8513 & 0.4673 & 0.9125 & 0.5396 & 0.8391 & 0.6250 & 0.3019 & 0.3190 \\
Gemini-3.1-Pro & \textbf{0.8374} & \textbf{0.7853} & 0.8761 & \textbf{0.6516} & 0.9190 & \textbf{0.7216} & \textbf{0.8704} & \textbf{0.8276} & 0.5556 & \textbf{0.5843} \\
Claude-Opus-4.6 & 0.7844 & 0.7338 & \textbf{0.8746} & 0.5407 & \textbf{0.9270} & 0.6057 & 0.8647 & 0.7208 & 0.3509 & 0.4240 \\
DeepSeek-V4-Pro & 0.8097 & 0.6922 & 0.8072 & 0.5075 & 0.8770 & 0.6412 & 0.8003 & 0.6667 & 0.5812 & 0.4399 \\
Qwen3.6-Max & 0.8238 & 0.7131 & 0.8283 & 0.5313 & 0.8949 & 0.6728 & 0.8212 & 0.7347 & \textbf{0.6098} & 0.4520 \\
GPT-OSS-20B & 0.3042 & 0.5723 & 0.8484 & 0.2323 & 0.9122 & 0.3010 & 0.8370 & 0.0196 & 0.1576 & 0.1145 \\
Qwen3.5-35B-A3B & 0.7362 & 0.6564 & 0.7903 & 0.4429 & 0.8700 & 0.5540 & 0.7827 & 0.5579 & 0.4062 & 0.3821 \\
\midrule
Average & 0.7109 & 0.6919 & 0.8395 & 0.4819 & 0.9018 & 0.5766 & 0.8308 & 0.5932 & 0.4233 & 0.3880 \\
\bottomrule
\end{tabular}%
}
\caption{Overall and class-level F1 of seven SOTA models on CAIL2018 charge prediction and PDP-Bench. CAIL Macro-F1 serves as the conventional LJP reference; L1 and L2 denote the binary and four-way PDP settings.}
\label{tab:ljp_pdp_transfer}
\end{table*}

\begin{table*}[!t]
\centering
\small
\renewcommand{\arraystretch}{1.08}
\resizebox{\textwidth}{!}{%
\begin{tabular}{@{}cc*{9}{c}@{}}
\toprule
\multirow{2}{*}{\textbf{Model}} & \multirow{2}{*}{\textbf{Budget}} & \multicolumn{4}{c}{\textbf{PDP L1}} & \multicolumn{5}{c}{\textbf{PDP L2}} \\
\cmidrule(lr){3-6}\cmidrule(l){7-11}
& & \textbf{Macro-F1} & \textbf{Micro-F1} & \textbf{NP} & \textbf{P} & \textbf{Macro-F1} & \textbf{Micro-F1} & \textbf{IENP} & \textbf{SNP} & \textbf{DNP} \\
\midrule
\multirow{5}{*}{GPT-5.4} & none & \textbf{0.7666} & \textbf{0.7956} & \textbf{0.8524} & \textbf{0.6809} & 0.5822 & 0.6042 & 0.6644 & 0.5355 & \textbf{0.4479} \\
& low & 0.7347 & 0.7593 & 0.8180 & 0.6514 & 0.6277 & 0.6425 & 0.7273 & 0.7455 & 0.3867 \\
& medium & 0.7161 & 0.7414 & 0.8050 & 0.6271 & 0.6395 & 0.6485 & \textbf{0.7712} & \textbf{0.7814} & 0.3784 \\
& high & 0.7291 & 0.7530 & 0.8106 & 0.6476 & \textbf{0.6399} & \textbf{0.6498} & 0.7554 & 0.7650 & 0.3915 \\
& xhigh & 0.4411 & 0.4772 & 0.5359 & 0.3463 & 0.4101 & 0.4129 & 0.4719 & 0.5870 & 0.2353 \\
\midrule
\multirow{3}{*}{Gemini-3.1-Pro} & low & 0.8284 & 0.8606 & 0.9032 & 0.7535 & 0.7815 & 0.7763 & 0.8793 & 0.8341 & 0.6589 \\
& medium & \textbf{0.8519} & \textbf{0.8785} & \textbf{0.9137} & \textbf{0.7900} & \textbf{0.8059} & \textbf{0.8036} & \textbf{0.8898} & 0.8393 & \textbf{0.7045} \\
& high & 0.8405 & 0.8651 & 0.8959 & 0.7852 & 0.8014 & 0.7992 & 0.8793 & \textbf{0.8496} & 0.6914 \\
\midrule
\multirow{4}{*}{Claude-Opus-4.6} & 0 & 0.8083 & 0.8337 & 0.8791 & 0.7375 & 0.6007 & 0.6212 & 0.7352 & 0.4615 & 0.4688 \\
& 1024 & \textbf{0.8189} & \textbf{0.8465} & \textbf{0.8911} & \textbf{0.7468} & \textbf{0.6766} & \textbf{0.6841} & \textbf{0.7597} & \textbf{0.6598} & \textbf{0.5401} \\
& 2048 & 0.8082 & 0.8373 & 0.8850 & 0.7314 & 0.6644 & 0.6727 & 0.7568 & 0.6316 & 0.5378 \\
& 4096 & 0.7923 & 0.8201 & 0.8698 & 0.7147 & 0.6525 & 0.6613 & 0.7529 & 0.6316 & 0.5108 \\
\midrule
\multirow{3}{*}{DeepSeek-V4-Pro} & none & \textbf{0.7754} & \textbf{0.8190} & \textbf{0.8769} & 0.6738 & 0.6946 & 0.6896 & \textbf{0.7196} & 0.7602 & 0.6250 \\
& high & 0.7513 & 0.7894 & 0.8492 & 0.6533 & 0.7056 & 0.7019 & 0.6897 & 0.8291 & 0.6504 \\
& xhigh & 0.7731 & 0.8110 & 0.8658 & \textbf{0.6804} & \textbf{0.7188} & \textbf{0.7175} & 0.6869 & \textbf{0.8333} & \textbf{0.6745} \\
\midrule
\multirow{5}{*}{Qwen3.6-Max} & 0 & 0.7716 & 0.7988 & 0.8499 & 0.6933 & 0.6344 & 0.6406 & 0.7094 & 0.5838 & 0.5512 \\
& 512 & \textbf{0.8584} & \textbf{0.8896} & \textbf{0.9258} & \textbf{0.7910} & 0.7141 & 0.7189 & 0.7448 & 0.6316 & \textbf{0.6890} \\
& 1024 & 0.8049 & 0.8402 & 0.8864 & 0.7234 & 0.7175 & 0.7136 & 0.7577 & 0.7488 & 0.6400 \\
& 2048 & 0.8160 & 0.8445 & 0.8870 & 0.7450 & \textbf{0.7435} & \textbf{0.7402} & 0.7876 & \textbf{0.8145} & 0.6270 \\
& 4096 & 0.8101 & 0.8420 & 0.8879 & 0.7322 & 0.7385 & 0.7340 & \textbf{0.8035} & 0.8091 & 0.6094 \\
\bottomrule
\end{tabular}%
}
\caption{PDP L1 and PDP L2 F1 of five SOTA models under different reasoning budgets. Values are means over eight repetitions.}
\label{tab:test_time_scaling}
\end{table*}

\section{Experiments}

We study PDP through three empirical questions. First, using trial-stage LJP as a reference, we test whether PDP challenges current SOTA LLMs (Section~\ref{sec:rq1}). Second, we examine whether test-time scaling, legal-domain specialization, or prompt-side knowledge augmentation can solve this challenge (Section~\ref{sec:rq2}). Finally, we run controlled RLVR interventions that increase the training share of each prosecution-decision class (Section~\ref{sec:rq3}).

We report Macro-F1, Micro-F1, and class-level F1 at two granularities: Level 1 (NP/P) and Level 2 (IENP/SNP/DNP/P); law-article citation F1 is in Appendix~\ref{sec:app_articles}. RQ1 uses one inference per sample, while RQ2 and RQ3 use eight independent inferences per sample unless noted otherwise. All evaluations are conducted on PDP-Bench (4{,}630 samples) without external retrieval, using the Baseline task instruction in Appendix~\ref{sec:app_prompts}. Model versions and decoding settings are in Appendix~\ref{sec:app_models}.

\subsection{Does PDP Challenge SOTA LLMs?}
\label{sec:rq1}

\paragraph{Setup.}
We evaluate seven models: three international closed models (GPT-5.4, Gemini-3.1-Pro, Claude-Opus-4.6), two Chinese closed models (DeepSeek-V4-Pro, Qwen3.6-Max), and two open-weight models (GPT-OSS-20B, Qwen3.5-35B-A3B). Each model performs one inference on PDP-Bench under its default API reasoning budget. As the LJP reference, we evaluate the same models on CAIL2018 charge prediction (data\_test\_charge\_4k, 4{,}000 samples) and report Macro-F1.

\begin{table*}[!t]
\centering
\small
\renewcommand{\arraystretch}{1.08}
\resizebox{\textwidth}{!}{%
\begin{tabular}{@{}cc*{9}{c}@{}}
\toprule
\multirow{2}{*}{\textbf{Paradigm}} & \multirow{2}{*}{\textbf{Model}} & \multicolumn{4}{c}{\textbf{PDP L1}} & \multicolumn{5}{c}{\textbf{PDP L2}} \\
\cmidrule(lr){3-6}\cmidrule(l){7-11}
& & \textbf{Macro-F1} & \textbf{Micro-F1} & \textbf{NP} & \textbf{P} & \textbf{Macro-F1} & \textbf{Micro-F1} & \textbf{IENP} & \textbf{SNP} & \textbf{DNP} \\
\midrule
\multirow{2}{*}{Full SFT} & Qwen2.5-7B-Instruct & \textbf{0.6271} & \textbf{0.7923} & \textbf{0.3816} & \textbf{0.8725} & \textbf{0.4160} & \textbf{0.7778} & \textbf{0.2937} & \textbf{0.1959} & \textbf{0.3020} \\
& LawLLM-7B & 0.3813 $\downarrow$ & 0.5886 $\downarrow$ & 0.1191 $\downarrow$ & 0.6435 $\downarrow$ & 0.2100 $\downarrow$ & 0.5843 $\downarrow$ & 0.0428 $\downarrow$ & 0.0664 $\downarrow$ & 0.0872 $\downarrow$ \\
\midrule
\multirow{2}{*}{\shortstack{R1-Distill\\SFT+GRPO}} & Qwen2.5-14B-Instruct & \textbf{0.5783} & \textbf{0.6658} & \textbf{0.3937} & 0.7628 & 0.4554 & \textbf{0.6514} & \textbf{0.4705} & 0.2578 & \textbf{0.3304} \\
& LegalDelta & 0.5582 $\downarrow$ & 0.6559 $\downarrow$ & 0.3520 $\downarrow$ & \textbf{0.7644} $\uparrow$ & \textbf{0.4644} $\uparrow$ & 0.6410 $\downarrow$ & 0.3964 $\downarrow$ & \textbf{0.4108} $\uparrow$ & 0.2860 $\downarrow$ \\
\midrule
\multirow{3}{*}{\shortstack{Mid-training+\\LEAD-SFT+RL}} & Qwen3-8B-Base & 0.5319 & 0.7406 & 0.2395 & 0.8244 & 0.2953 & 0.7313 & 0.0597 & 0.0964 & 0.2006 \\
& Qwen3-8B Instruct & \textbf{0.6163} $\uparrow$ & \textbf{0.8657} $\uparrow$ & 0.3070 $\uparrow$ & \textbf{0.9255} $\uparrow$ & 0.4551 $\uparrow$ & \textbf{0.8601} $\uparrow$ & 0.3448 $\uparrow$ & 0.3277 $\uparrow$ & 0.2224 $\uparrow$ \\
& LegalOne-R1-8B & 0.6124 $\downarrow$ & 0.7277 $\downarrow$ & \textbf{0.4170} $\uparrow$ & 0.8079 $\downarrow$ & \textbf{0.5391} $\uparrow$ & 0.7213 $\downarrow$ & \textbf{0.4212} $\uparrow$ & \textbf{0.5524} $\uparrow$ & \textbf{0.3749} $\uparrow$ \\
\bottomrule
\end{tabular}%
}
\caption{Overall and class-level F1 of three base-legal model pairs on PDP-Bench. Values are means over eight repetitions. For Mid-training+LEAD-SFT+RL, we additionally report Qwen3-8B Instruct as a fairer baseline. Bold marks the better value within each comparison block, and arrows indicate directional changes relative to the preceding row.}
\label{tab:legal_specialization}
\end{table*}

\begin{table*}[!t]
\centering
\small
\renewcommand{\arraystretch}{1.08}
\resizebox{\textwidth}{!}{%
\begin{tabular}{@{}cc*{9}{c}@{}}
\toprule
\multirow{2}{*}{\textbf{Model}} & \multirow{2}{*}{\textbf{Prompt}} & \multicolumn{4}{c}{\textbf{PDP L1}} & \multicolumn{5}{c}{\textbf{PDP L2}} \\
\cmidrule(lr){3-6}\cmidrule(l){7-11}
& & \textbf{Macro-F1} & \textbf{Micro-F1} & \textbf{NP} & \textbf{P} & \textbf{Macro-F1} & \textbf{Micro-F1} & \textbf{IENP} & \textbf{SNP} & \textbf{DNP} \\
\midrule
\multirow{3}{*}{GPT-5.4} & Baseline & 0.7634 & 0.7892 & 0.8429 & 0.6838 & 0.6006 & 0.6180 & 0.6853 & 0.5699 & \textbf{0.4635} \\
& +Definitions & 0.7612 & 0.7900 & 0.8466 & 0.6758 & \textbf{0.6220} & \textbf{0.6453} & \textbf{0.7665} & \textbf{0.6317} & 0.4142 \\
& +One-shot & \textbf{0.7724} & \textbf{0.7972} & \textbf{0.8497} & \textbf{0.6952} & 0.5919 & 0.6177 & 0.7178 & 0.4973 & 0.4575 \\
\midrule
\multirow{3}{*}{Gemini-3.1-Pro} & Baseline & 0.8146 & 0.8426 & 0.8783 & 0.7509 & 0.7801 & 0.7779 & 0.8572 & \textbf{0.8530} & 0.6594 \\
& +Definitions & 0.8220 & 0.8572 & 0.8990 & 0.7449 & 0.7902 & 0.7877 & \textbf{0.8839} & 0.8433 & 0.6887 \\
& +One-shot & \textbf{0.8642} & \textbf{0.8894} & \textbf{0.9201} & \textbf{0.8082} & \textbf{0.8089} & \textbf{0.8081} & 0.8808 & 0.8220 & \textbf{0.7246} \\
\midrule
\multirow{3}{*}{Claude-Opus-4.6} & Baseline & \textbf{0.8063} & \textbf{0.8325} & \textbf{0.8775} & \textbf{0.7352} & 0.5814 & 0.6012 & 0.7233 & 0.4396 & 0.4276 \\
& +Definitions & 0.8035 & 0.8300 & 0.8757 & 0.7313 & \textbf{0.6617} & \textbf{0.6737} & \textbf{0.8039} & \textbf{0.6036} & \textbf{0.5078} \\
& +One-shot & 0.7891 & 0.8170 & 0.8671 & 0.7111 & 0.6285 & 0.6403 & 0.7608 & 0.5441 & 0.4982 \\
\midrule
\multirow{3}{*}{DeepSeek-V4-Pro} & Baseline & 0.7567 & 0.7914 & 0.8494 & 0.6639 & 0.7057 & 0.7039 & 0.7006 & \textbf{0.8456} & 0.6128 \\
& +Definitions & 0.7127 & 0.7430 & 0.8048 & 0.6205 & 0.6363 & 0.6348 & 0.6592 & 0.7402 & 0.5254 \\
& +One-shot & \textbf{0.7809} & \textbf{0.8241} & \textbf{0.8798} & \textbf{0.6819} & \textbf{0.7199} & \textbf{0.7162} & \textbf{0.7224} & 0.7911 & \textbf{0.6842} \\
\midrule
\multirow{3}{*}{Qwen3.6-Max} & Baseline & 0.7901 & 0.8192 & 0.8678 & 0.7123 & 0.7026 & 0.7004 & 0.7631 & 0.7809 & 0.5542 \\
& +Definitions & 0.8059 & 0.8405 & 0.8875 & 0.7242 & 0.7403 & 0.7342 & 0.8140 & \textbf{0.8074} & 0.6158 \\
& +One-shot & \textbf{0.8529} & \textbf{0.8825} & \textbf{0.9188} & \textbf{0.7871} & \textbf{0.7691} & \textbf{0.7675} & \textbf{0.8340} & 0.7806 & \textbf{0.6749} \\
\bottomrule
\end{tabular}%
}
\caption{PDP L1 and PDP L2 F1 of five SOTA models under Baseline, +Definitions, and +One-shot prompt conditions. Values are means over eight repetitions.}
\label{tab:prompt_augmentation}
\end{table*}

\paragraph{Analysis.}
Table~\ref{tab:ljp_pdp_transfer} shows a clear transfer gap from LJP to PDP: although CAIL2018 has more than 200 charge labels and PDP L2 has only four, the five closed models lose roughly 0.12--0.18 Macro-F1 from CAIL to PDP L2, with a further 0.04--0.15 drop from L1 to L2 (open-weight models follow the same pattern); only Gemini-3.1-Pro exceeds 0.7 on L2. Class-level scores expose the hardest boundaries: the best F1 reaches 0.83 on IENP and 0.93 on P, but only 0.61 on SNP and 0.58 on DNP. Model-level correlations between LJP and PDP class metrics are in Appendix~\ref{app:ljp_pdp_correlation}.


\paragraph{Finding 1: PDP is a systematic challenge concentrated on legal subsumption and value-based discretion.}
The PDP gap is structurally allocated to SNP and DNP, the two boundaries that correspond to legal subsumption and value-based discretion, rather than uniformly distributed across classes. We next ask whether mainstream enhancement routes can close it.

\subsection{Can Mainstream Enhancement Help?}
\label{sec:rq2}

We test three common routes: test-time scaling, legal-domain specialization, and prompt-side knowledge augmentation. A route counts as solving the PDP challenge only if it simultaneously satisfies two criteria: (a) consistent gains across models and classes, and (b) removal of the SNP/DNP ceiling identified in Finding~1. We check each route against both criteria below.

\subsubsection{Test-Time Scaling}
\label{sec:test_time_scaling}

\paragraph{Setup.}
We use five representative SOTA models from Section~\ref{sec:rq1} and vary only the inference budget. Interventions use each model's native OpenRouter interface, either effort or max\_tokens; valid levels are listed in Appendix~\ref{sec:app_budget_levels}. Evaluation uses \emph{test\_rq2}, a 100-sample subset of PDP-Bench formed by uniformly sampling 25 instances per decision class for low-budget multi-repeat diagnostics; each sample is inferred 8 times independently, yielding 800 inferences per budget level. Full estimates with standard errors are in Appendix~\ref{sec:app_budget_full}.

\paragraph{Analysis.}
As shown in Table~\ref{tab:test_time_scaling}, test-time scaling is non-monotonic and model-specific: Gemini-3.1-Pro peaks at medium, Claude-Opus-4.6 at 1024 then declines, Qwen3.6-Max at 512 for L1 but 2048 for L2, and GPT-5.4 collapses at xhigh while obtaining its best L1 score with no added budget. At the class level, longer inference often helps IENP and SNP (e.g., GPT-5.4 on SNP: $0.54{\to}0.78$; DeepSeek-V4-Pro: $0.76{\to}0.83$), but DNP responds contradictorily across models: GPT-5.4 degrades to 0.24 at xhigh, DeepSeek-V4-Pro rises to 0.67, and the others peak at low or medium budgets. Test-time scaling thus violates both criteria: (a) the optimal budget is model-specific, and (b) the best DNP F1 across budgets reaches only 0.7045.

\begin{figure*}[!t]
    \centering
    \includegraphics[width=\textwidth]{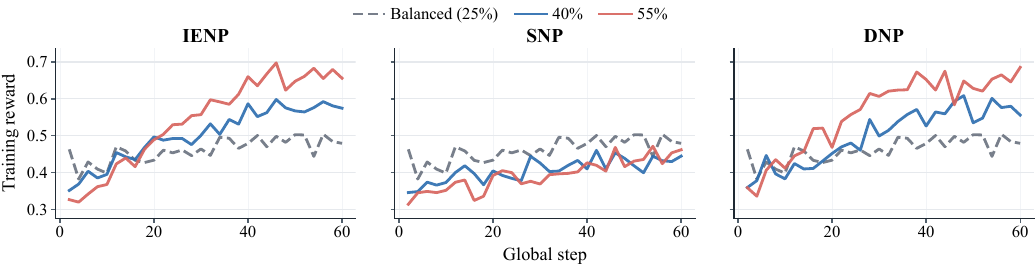}
    \caption{DAPO training reward curves under IENP, SNP, and DNP target-class interventions. The x-axis is global step; the three curves denote balanced 25\%, moderate 40\%, and controlled strong 55\% target-class training proportions.}
    \label{fig:rq3_reward_curves}
\end{figure*}

\begin{figure*}[!t]
    \centering
    \includegraphics[width=\textwidth]{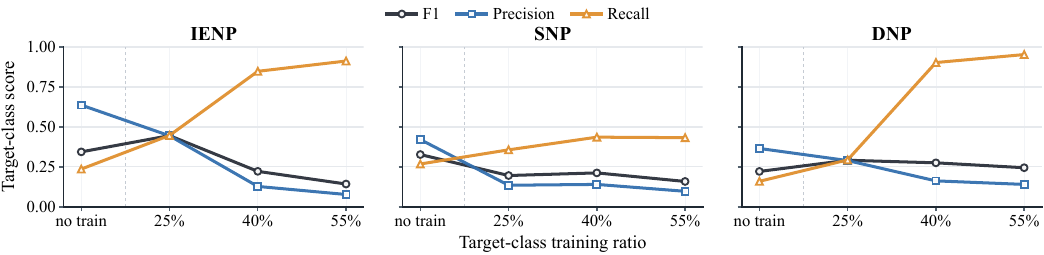}
    \caption{Target-class F1, precision, and recall on PDP-Bench under IENP, SNP, and DNP class-prior interventions. The x-axis denotes the target-class training proportion (no train / 25\% / 40\% / 55\%), where ``no train'' reports the un-tuned Qwen3-8B baseline.}
    \label{fig:rq3_class_prior}
\end{figure*}

\subsubsection{Legal-Domain Specialization}
\label{sec:legal_specialization}

\paragraph{Setup.}
We compare matched base and legal-domain models. Within each pair, scale and upstream base are fixed; the key difference is legal-domain specialization. The pairs cover Full SFT (DISC-LawLLM; \citep{disc-lawllm}), R1-Distill SFT+GRPO (LegalDelta; \citep{legaldelta}), and Mid-training+LEAD-SFT+RL (LegalOne; \citep{legalone}). Evaluation uses PDP-Bench; overall scores with standard errors are in Appendix~\ref{sec:app_legal_full}.

\paragraph{Analysis.}
Table~\ref{tab:legal_specialization} shows that the three paradigms transfer differently. Full SFT hurts across all metrics: LawLLM-7B falls below 0.10 F1 on all three L2 NP subclasses, suggesting a mismatch between common legal-SFT data and prosecutorial review. R1-Distill SFT+GRPO is mostly neutral: LegalDelta improves SNP but loses on L1, IENP, and DNP. Mid-training+LEAD-SFT+RL is the only recipe that consistently improves all three NP classes, but its gain is partly instruction following: Qwen3-8B-Base to Qwen3-8B (Instruct) already raises L2 Macro-F1 from 0.2953 to 0.4551, and LegalOne adds only +0.0840 over the Instruct baseline while L1 P drops from 0.9255 to 0.8079---a shift toward NP at the cost of P stability. Legal-domain specialization thus violates both criteria: (a) gains depend on one specific recipe, and (b) even this best recipe attains only 0.5524/0.3749 on SNP/DNP.

\subsubsection{Prompt-Side Knowledge Augmentation}
\label{sec:prompt_aug}

\paragraph{Setup.}
We test three prompt conditions: Baseline, +Definitions, and +One-shot. +Definitions adds legal definitions of the four decisions; +One-shot adds one representative case per decision. Models and data follow Section~\ref{sec:test_time_scaling}, and reasoning budgets use API defaults. The full prompts for the three conditions are shown in Appendix~\ref{sec:app_prompts}; full estimates with standard errors are in Appendix~\ref{sec:app_prompt_full}.

\paragraph{Analysis.}
As shown in Table~\ref{tab:prompt_augmentation}, both +Definitions and +One-shot improve L2 Macro-F1 for four of five models, but model preferences split: two prefer +Definitions and three prefer +One-shot. The class effects are uneven: on DNP, +One-shot helps four models by +0.065 to +0.121, while +Definitions helps three models but hurts two; on SNP, despite a maximum gain of +0.1640, two of five models regress under the prompt that maximizes their own L2 Macro-F1. Prompt augmentation thus violates both criteria: (a) the best prompt is model-specific, and (b) the best DNP F1 across prompt conditions reaches only 0.7246.

\paragraph{Finding 2: Mainstream enhancement routes give only local gains and do not remove the SNP/DNP ceiling.}
All three routes violate both criteria, so the SNP/DNP ceiling identified in Finding 1 remains. This motivates our final question: can class-augmented RLVR partially resolve PDP?

\subsection{Can simple outcome-reward RLVR Help?}
\label{sec:rq3}

\paragraph{Setup.}
We conduct controlled class-prior interventions with Qwen3-8B and DAPO~\citep{yu2026dapo}, fixing the algorithm, training budget, prompt, and evaluation set while varying only the target-class share. The seven \emph{pdp2k\_rq3} training splits each contain 2{,}000 samples drawn from a prosecutorial-document pool that is fully disjoint from PDP-Bench: a balanced 25\% baseline and 40\%/55\% interventions for IENP, SNP, and DNP. Rewards depend only on output format and whether the predicted decision matches the gold label; this is the \emph{simple binary correctness reward} that dominates current RLVR practice. Evaluation uses PDP-Bench; full split counts, hyperparameters, and data-source details are in Appendix~\ref{sec:app_rq3_config}.

\paragraph{Analysis.}
Figures~\ref{fig:rq3_reward_curves} and~\ref{fig:rq3_class_prior} report training reward curves and test-time target-class F1/precision/recall; ``no train'' marks the un-tuned Qwen3-8B baseline (full results in Appendix~\ref{sec:app_rq3_full}). For IENP and DNP, training reward rises with the target-class share (reaching about 0.65--0.70 near step 60 under 55\%), but on the test set this becomes recall inflation with collapsing precision: from no-train to 55\%, IENP precision $0.64{\to}0.08$ and recall $0.24{\to}0.91$ drag F1 from 0.34 down to 0.14; DNP precision $0.37{\to}0.14$ and recall $0.16{\to}0.95$ keep F1 stuck at 0.22--0.29 throughout. SNP behaves differently and more harmfully: reward curves nearly overlap around 0.45, and from the baseline (P=0.42, R=0.27, F1=0.33) training only modestly lifts recall to 0.43 while collapsing precision to 0.10, so F1 \emph{degrades} to 0.16---roughly half the baseline.

A simple binary correctness reward feeds back \emph{whether the output label is right}, but PDP evaluates \emph{whether the discriminative conditions hold}---evidence sufficiency, legal subsumption, and value-based discretion. This outcome-versus-process mismatch surfaces as two complementary pathologies on the three NP boundaries.

\textit{Prior-amplification shortcut} (IENP, DNP). The base model's pre-RL prior already produces enough target-class rollouts for binary rewards to fire. Class augmentation reweights these positive rollouts upward, and the policy can satisfy the reward by simply emitting the target label more often---without acquiring the underlying evidence-sufficiency or discretionary-circumstance criteria. RLVR thus amplifies a Bayesian prior into a label-output shortcut, not a discriminator, matching the observed recall surge with collapsing precision.

\textit{Sparse-signal dead zone} (SNP). Statutory non-prosecution requires recognizing categorical legal exemptions (e.g., self-defense, statute-of-limitations, criminal-responsibility age) that the base model rarely emits unprompted. Near-zero target-class rollouts means near-zero positive signal under a binary outcome reward, so the policy receives little gradient toward discrimination regardless of the training share---matching the overlapping reward curves---and the small movement that does occur is a precision collapse that pushes SNP F1 below its un-tuned baseline.

Class augmentation thus reshapes \emph{which} labels the model is rewarded for, not \emph{what} it is rewarded for: a sparse hit/miss that can be satisfied by prior amplification but not by genuine discrimination.

\paragraph{Finding 3: Class-augmented RLVR with a simple binary correctness reward does not produce generalizable PDP discrimination.}
The SNP/DNP ceiling identified in Findings 1--2 remains open under this reward design. Closing it likely calls for process-level or rationale-grounded reward signals that target PDP's three capabilities directly, rather than further reweighting of the same binary signal.

%% file: sections/5_Conclusion.tex
\section{Conclusion}

Trial-stage LJP sees only indicted cases and therefore overlooks three of the four core criminal-liability outcomes. To fill this gap, we propose \textbf{Prosecution Decision Prediction (PDP)}---a four-way prosecutorial-stage task aligned with evidence evaluation, legal subsumption, and value-based discretion---and construct \textbf{PDP-Bench}, a benchmark of 4{,}630 publicly available Chinese prosecutorial decisions spanning 190 charges.

Three findings together establish PDP as an open challenge for current Legal AI. First, SOTA LLMs lose 0.12--0.18 Macro-F1 from CAIL2018 charge prediction to PDP-Bench, with the gap structurally concentrated on SNP and DNP. Second, none of the three mainstream enhancement routes---test-time scaling, legal-domain specialization, and prompt-side knowledge augmentation---delivers consistent cross-model, cross-class gains or removes this SNP/DNP ceiling. Third, class-augmented RLVR under a simple binary correctness reward only reshapes label preferences---amplifying priors on IENP/DNP and leaving the SNP boundary with almost no learning signal---so target-class F1 fails to improve and even degrades. PDP-Bench therefore offers a new evaluation anchor for criminal-liability reasoning, pointing to process-level or rationale-grounded rewards as a promising direction for closing the remaining gap.

%% file: sections/Limitations.tex
\section*{Limitations}
While our work introduces the PDP task and provides a comprehensive evaluation of SOTA LLMs on Chinese criminal data, several limitations should be acknowledged.
 
\paragraph{Jurisdictional Scope of Data.} Prosecutorial review is a common legal procedure across most major jurisdictions, which means the PDP task itself is a general legal AI problem rather than one confined to the Chinese judicial system. 
However, due to the inherent difficulty of collecting large-scale legal data across jurisdictions, our PDP-Bench is constructed solely from Chinese prosecutorial review documents. 
We leave the construction of cross-jurisdictional PDP benchmarks to future work.
 
\paragraph{Class Imbalance in PDP-Bench.} In Chinese judicial practice, non-prosecution decisions are released to the public far less frequently than prosecution decisions. 
This natural skew leads to a highly imbalanced label distribution in PDP-Bench. 
Consequently, when interpreting evaluation results, readers should place greater emphasis on per-class metrics and macro-averaged scores rather than relying solely on overall accuracy, especially for the minority non-prosecution categories where overall accuracy can be misleading.

\section*{Ethical Considerations}
Although PDP-Bench contains criminal case data, we have removed personally identifying information, so it does not compromise individual privacy. Furthermore, the PDP task itself is designed to help lawyers and criminal suspects anticipate potential prosecutorial review outcomes and to enhance judicial transparency; it is not intended to replace or automate the existing judicial system. We therefore believe that this work raises no substantive ethical concerns.

%% file: sections/Appendix.tex
\section{Model Versions and API Endpoints}
\label{sec:app_models}

The models used in Sections~\ref{sec:rq1}, \ref{sec:test_time_scaling}, and \ref{sec:prompt_aug} are accessed through the OpenRouter API. Table~\ref{tab:app_openrouter_models} lists the endpoints used during the evaluation period from February to April 2026.

\begin{table}[!htbp]
\centering
\footnotesize
\setlength{\tabcolsep}{3pt}
\resizebox{\columnwidth}{!}{%
\begin{tabular}{@{}ll@{}}
\toprule
\textbf{Model} & \textbf{OpenRouter Endpoint} \\
\midrule
GPT-5.4\citep{openai2026gpt54} & \texttt{openai/gpt-5.4} \\
Gemini-3.1-Pro\citep{google2026gemini31pro} & \texttt{google/gemini-3.1-pro-preview} \\
Claude-Opus-4.6\citep{anthropic2026claudeopus46} & \texttt{anthropic/claude-opus-4.6} \\
DeepSeek-V4-Pro\citep{deepseek2026v4pro} & \texttt{deepseek/deepseek-v4-pro} \\
Qwen3.6-Max\citep{qwen2026qwen36max} & \texttt{qwen/qwen3.6-max-preview} \\
GPT-OSS-20B\citep{openai2025gptoss} & \texttt{openai/gpt-oss-20b} \\
Qwen3.5-35B-A3B\citep{qwen2026qwen35a3b} & \texttt{qwen/qwen3.5-35b-a3b} \\
\bottomrule
\end{tabular}%
}
\caption{OpenRouter endpoints used in the evaluation.}
\label{tab:app_openrouter_models}
\end{table}

The three base--legal model pairs in Section~\ref{sec:legal_specialization} are open-weight models loaded from HuggingFace and evaluated locally with vLLM. Table~\ref{tab:app_hf_models} lists the repositories.

\begin{table}[!htbp]
\centering
\footnotesize
\setlength{\tabcolsep}{3pt}
\resizebox{\columnwidth}{!}{%
\begin{tabular}{@{}ll@{}}
\toprule
\textbf{Model} & \textbf{HuggingFace Repository} \\
\midrule
Qwen2.5-7B-Instruct\cite{qwen2024qwen25} & \texttt{Qwen/Qwen2.5-7B-Instruct} \\
LawLLM-7B\citep{disc-lawllm} & \texttt{ShengbinYue/LawLLM-7B} \\
Qwen3-8B-Base\cite{qwen2025qwen3} & \texttt{Qwen/Qwen3-8B-Base} \\
Qwen3-8B\cite{qwen2025qwen3} & \texttt{Qwen/Qwen3-8B} \\
LegalOne-R1-8B\citep{legalone} & \texttt{LegalAI/LegalOne-R1-8B} \\
Qwen2.5-14B-Instruct\citep{qwen2024qwen25} & \texttt{Qwen/Qwen2.5-14B-Instruct} \\
LegalDelta\citep{legaldelta} & \texttt{LegalAI/LegalDelta} \\
\bottomrule
\end{tabular}%
}
\caption{HuggingFace repositories for open-weight models.}
\label{tab:app_hf_models}
\end{table}

Except for the reasoning-budget interventions in Section~\ref{sec:test_time_scaling}, OpenRouter calls use \texttt{max\_tokens=8192}; \texttt{temperature} and \texttt{top\_p} follow each model API's defaults. The RQ1 evaluation in Section~\ref{sec:rq1} uses one inference per sample on PDP-Bench. The repeated-run experiments in Sections~\ref{sec:test_time_scaling}, \ref{sec:legal_specialization}, \ref{sec:prompt_aug}, and~\ref{sec:rq3} use 8 independent inference runs with \texttt{seed = base\_seed + run\_id} and \texttt{base\_seed = 42}. The vLLM evaluations in Section~\ref{sec:legal_specialization} are run locally on a single 4090 48GB GPU with default decoding parameters.

\section{Data Construction Prompts}
\label{sec:app_data_prompts}

This appendix section gives English renderings of the implementation prompts used for dual-track extraction, extraction review, and sensitive-information de-identification in Section~\ref{sec:pdp}. Stage 3 data refinement is described in the main text and does not use a separate prompt reported here. We keep the original role split, field order, required output schemas, and line-level organization; runtime variables such as \{raw\_text\} and \{record\_json\} are retained as placeholders.

\subsection{Dual-Track Extraction and Review}

\paragraph{Extraction prompt.}
The following prompt is used by the LLM extraction track in parallel with the rule-based extractor.

\begin{figure*}[!t]
\begin{tcolorbox}[enhanced jigsaw, title=Document Extraction Prompt: System and One-Shot Input, colback=white, colframe=black!35, boxrule=0.4pt, arc=1mm, left=1mm, right=1mm, top=1mm, bottom=1mm]
\footnotesize
\textbf{System Prompt}\\
You are an assistant for structured extraction and quality control of Chinese criminal prosecutorial documents.\\
Your task is to clean raw prosecutorial documents into high-quality training data.\\
\textbf{Requirements:}\\
1. Be faithful to the original text. Do not fabricate facts, law articles, times, places, relationships, or dispositions that do not appear in the document.\\
2. You may think, compare, locate paragraphs, and self-check internally, but the final answer must contain only one valid JSON object.\\
3. Do not output reasoning, explanations, markdown, prefixes, suffixes, or any extra text in the final answer.\\
4. If uncertain, use an empty string or an empty array. Do not guess.\\

- - -\\
\textbf{User Prompt: One-Shot Example Input}\\
{}\textbf{[Example input directory type]} Prosecution\\
{}\textbf{[Example input case number]} Qi Xin Procuratorial Criminal Indictment [2024] No. 74\\
{}\textbf{[Example input full text]} The example is a public indictment for a dangerous-driving case. It contains a webpage source line, the procuratorate name, the indictment title and case number, defendant identity information, case-transfer and rights-notification procedures, the reviewed facts, an evidence catalogue, the prosecutorial reasoning, the dispositive request, court submission line, prosecutor signatures, date, and attachments. The facts state that the defendant drove an unlicensed three-wheeled motorcycle while intoxicated and was seized by traffic police; the evidence catalogue includes documentary evidence, the defendant's confession and defense, and expert opinion.
\end{tcolorbox}
\caption{System instruction and one-shot input template for document extraction.}
\label{fig:app_extract_system_input}
\end{figure*}

\begin{figure*}[!t]
\begin{tcolorbox}[enhanced jigsaw, title=Document Extraction Prompt: One-Shot Output and Rules, colback=white, colframe=black!35, boxrule=0.4pt, arc=1mm, left=1mm, right=1mm, top=1mm, bottom=1mm]
\footnotesize
{}\textbf{[Example output]}\\
\{``id'':``Qi Xin Procuratorial Criminal Indictment [2024] No. 74'', ``meta'':\{``date'':``2024-09-10'', ``province'':``Heilongjiang Province''\}, ``person\_info'':``The suspect Lu ..., released on bail pending trial for suspected dangerous driving.'', ``procedure'':``The case was investigated and concluded ..., rights were notified, opinions were heard, and all case materials were reviewed.'', ``fact'':``At around 9:30 on January 10, 2024, the suspect drove an unlicensed three-wheeled motorcycle while intoxicated ... The evidence proving the above facts is as follows: 1. documentary evidence; 2. the suspect's confession and defense; 3. expert opinion.'', ``relevant\_articles'':[``cl:Article 133-1'', ``cpl:Article 176''], ``decision'':``prosecution'', ``raw\_reasoning\_and\_decision'':``This Procuratorate holds that ... files a public prosecution and requests sentencing according to law.'', ``source\_url'':``''\}\\
{}\textbf{[Example points]}\\
1. Convert \emph{person\_info}, \emph{procedure}, and \emph{fact} uniformly to ``suspect''.\\
2. \emph{fact} must retain concrete conduct facts; immediately following investigation supplements and the evidence catalogue may be retained, but it cannot contain only an evidence catalogue.\\
3. If the document contains multiple fact versions, keep the last version, namely the version confirmed by this Procuratorate.\\
4. \emph{fact} must not contain reasoning or conclusions such as ``this Procuratorate holds'', ``according to'', ``decision as follows'', ``file a public prosecution'', or ``decide not to prosecute''.\\
5. \emph{raw\_reasoning\_and\_decision} keeps only the reasoning and dispositive text after ``this Procuratorate holds'', excluding submission text, attachments, signatures, and dates.\\
6. \emph{relevant\_articles} keeps only law articles that actually appear in the reasoning.
\end{tcolorbox}
\caption{One-shot output example and extraction rules for document extraction.}
\label{fig:app_extract_output_rules}
\end{figure*}

\begin{figure*}[!t]
\begin{tcolorbox}[enhanced jigsaw, title=Document Extraction Prompt: Actual Input, colback=white, colframe=black!35, boxrule=0.4pt, arc=1mm, left=1mm, right=1mm, top=1mm, bottom=1mm]
\footnotesize
Please process the following prosecutorial document and output only one valid JSON object. The keys must be complete and exactly match the following structure:\\
\{``id'':``'', ``meta'':\{``date'':``'', ``province'':``''\}, ``person\_info'':``'', ``procedure'':``'', ``fact'':``'', ``relevant\_articles'':[], ``decision'':``'', ``raw\_reasoning\_and\_decision'':``'', ``source\_url'':``''\}\\
{}\textbf{[Current task]}\\
1. If the actual document is an indictment, \emph{decision} must be ``prosecution''.\\
2. If the actual document is a non-prosecution decision, infer whether \emph{decision} is ``discretionary non-prosecution'', ``statutory non-prosecution'', or ``insufficient-evidence non-prosecution''.\\
3. \emph{person\_info}, \emph{procedure}, and \emph{fact} must uniformly use ``suspect''; do not retain ``defendant'' or ``non-prosecuted person''.\\
4. \emph{person\_info} keeps only identity, residence, coercive measures, and similar personal information.\\
5. \emph{procedure} keeps only investigation, transfer for prosecutorial review, supplementary investigation, extension, rights notification, interrogation, opinion hearing, and related procedural information.\\
6. \emph{fact} must retain concrete conduct facts; if a supplementary ``upon investigation'' narrative or evidence catalogue follows the fact section, it may be retained.\\
7. If several fact versions appear, keep the last one, namely the version confirmed by procuratorial review.\\
8. \emph{fact} must not contain only an evidence catalogue, evidence evaluation, or summary such as ``the above evidence was collected lawfully'', and must not mix in reasoning or conclusions.\\
9. \emph{raw\_reasoning\_and\_decision} keeps only the reasoning and dispositive text after ``this Procuratorate holds'', excluding title, source line, submission text, attachments, signatures, and date.\\
10. \emph{relevant\_articles} keeps only articles that actually appear in \emph{raw\_reasoning\_and\_decision}, with prefixes \emph{cl:/cpl:/cpr:}.\\
11. Normalize \emph{meta.date} to YYYY-MM-DD and \emph{meta.province} to the full provincial-level administrative-region name.\\
12. Delete webpage noise such as source, author, editor, time, font, and HTML fragments.\\
13. Be faithful to the original. If uncertain, leave blank. Do not fabricate.\\
{}\textbf{[File-name case number]} \{stem\_id\}\\
{}\textbf{[Original full document]} \{raw\_text\}
\end{tcolorbox}
\caption{Actual input template used by the LLM extraction track.}
\label{fig:app_extract_actual_input}
\end{figure*}

\paragraph{Review prompt.}
The review prompt is used when programmatic validation detects empty fields, label errors, fact--reasoning contamination, label-leaking appellations, or inconsistent law articles.

\begin{figure*}[!t]
\begin{tcolorbox}[enhanced jigsaw, title=Extraction Review Prompt, colback=white, colframe=black!35, boxrule=0.4pt, arc=1mm, left=1mm, right=1mm, top=1mm, bottom=1mm]
\footnotesize
\textbf{System Prompt}\\
You are a quality-control and revision assistant for Chinese criminal prosecutorial-document data.\\
You will see the original document, the current JSON, and a list of detected issues.\\
Please revise the current JSON according to the original document. Output only one valid JSON object and no explanation.\\
The revision must be faithful to the original and must not fabricate content.\\

- - -\\
\textbf{User Prompt}\\
Please revise the extraction result below and output only the revised JSON object.\\
{}\textbf{[Directory-type hint]} \{doc\_type\_cn\}\\
{}\textbf{[File-name case number]} \{stem\_id\}\\
{}\textbf{[Current issues]} \{issues\_text\}\\
{}\textbf{[Current JSON]} \{current\_json\}\\
{}\textbf{[Rule-extraction reference]} \{fallback\_json\}\\
{}\textbf{[Revision requirements]}\\
1. Preserve the original schema. Do not add or delete keys.\\
2. Prioritize fixing empty fields, decision errors, fact--reasoning mixing, label-leaking appellations, and missing or inconsistent law articles.\\
3. \emph{person\_info}, \emph{procedure}, and \emph{fact} must use the neutral term ``suspect''.\\
4. \emph{fact} must retain concrete conduct facts; investigation supplements and the evidence catalogue may be retained, but it cannot contain only an evidence catalogue or evidence evaluation.\\
5. If the fact section contains multiple versions, keep the last one, namely the version confirmed by this Procuratorate; do not mix in reasoning conclusions.\\
6. \emph{raw\_reasoning\_and\_decision} keeps only reasoning and dispositive text, excluding submission text, attachments, signatures, dates, and webpage noise.\\
7. \emph{relevant\_articles} keeps only law articles that truly appear in the reasoning.\\
8. Be faithful to the original. If uncertain, leave blank.\\
{}\textbf{[Original full document]} \{raw\_text\}
\end{tcolorbox}
\caption{Prompt used to review and revise extraction outputs flagged by validation.}
\label{fig:app_extract_review_prompt}
\end{figure*}

\subsection{Sensitive-Information De-Identification}

\paragraph{De-identification prompt.}
The de-identification prompt applies minimally necessary masking while preserving legally relevant facts.

\begin{figure*}[!t]
\begin{tcolorbox}[enhanced jigsaw, title=Sensitive-Information De-Identification Prompt: System, colback=white, colframe=black!35, boxrule=0.4pt, arc=1mm, left=1mm, right=1mm, top=1mm, bottom=1mm]
\footnotesize
\textbf{System Prompt}\\
You are a Chinese criminal prosecutorial-document de-identification assistant, specialized in personal sensitive information in public legal documents.\\
Task: apply minimally necessary de-identification to structured prosecutorial-document records. You must preserve legal facts, narrative order, evidence structure, charges, amounts, quantities, incident times, procedural times, law articles, and dispositions as much as possible, replacing only sensitive information that can identify natural persons or private entities.\\
You may conduct internal thinking, list mappings, and self-check. If you output the thinking process, it must appear before the final JSON. The last paragraph of the final answer must be one standalone complete JSON object parseable by \emph{json.loads}; no text may follow the JSON. Do not wrap the JSON in a markdown code block.\\
\textbf{Core principles:}\\
1. Be faithful to the original. Do not add facts or delete conduct, amounts, times, evidence names, or law articles that affect factual judgment.\\
2. Within the same record, the same natural person must use a consistent de-identified name; existing de-identified names may be preserved.\\
3. Real names of suspects, victims, witnesses, accomplices, relatives, vehicle owners, recipients, bank-card holders, and other natural persons must be de-identified.\\
4. Public institutions and public places such as public security organs, procuratorates, courts, forensic centers, administrative agencies, roads, expressways, and toll stations generally should not be de-identified.\\
5. Private addresses, precise registered-residence house numbers, delivery addresses, private companies, shops, training schools, residential compounds, and other information that can identify a person or private entity must be de-identified.\\
6. Return only four keys: \emph{person\_info}, \emph{procedure}, \emph{fact}, and \emph{raw\_reasoning\_and\_decision}; do not add, delete, or rename keys.
\end{tcolorbox}
\caption{System instruction for sensitive-information de-identification.}
\label{fig:app_deid_system_prompt}
\end{figure*}

\begin{figure*}[!t]
\begin{tcolorbox}[enhanced jigsaw, title=Sensitive-Information De-Identification Prompt: User, colback=white, colframe=black!35, boxrule=0.4pt, arc=1mm, left=1mm, right=1mm, top=1mm, bottom=1mm]
\footnotesize
Please de-identify sensitive information in the text fields of the following JSON record. You may think first, but the final output's last paragraph must be a single JSON object: \{``person\_info'':``'', ``procedure'':``'', ``fact'':``'', ``raw\_reasoning\_and\_decision'':``''\}.\\
{}\textbf{[Input field descriptions]}\\
-- \emph{person\_info}: identity, birth date, ID number, registered/current address, workplace, coercive measures, prior offenses, etc.\\
-- \emph{procedure}: investigation, transfer for prosecutorial review, supplementary investigation, extension, rights notification, interrogation, opinion hearing, etc.\\
-- \emph{fact}: case facts and evidence catalogue, often containing suspects, victims, witnesses, vehicles, bank cards, phone numbers, and delivery addresses.\\
-- \emph{raw\_reasoning\_and\_decision}: reasoning, law-article application, and disposition; it may also repeat names of natural persons.\\
{}\textbf{[Sensitive information that must be processed]}\\
1. Names of natural persons, including suspects, victims, witnesses, accomplices, recipients, relatives, vehicle owners, and bank-card holders. Two-character Chinese names are usually de-identified as surname + ``Mou''; names with three or more characters as surname + ``Moumou''; multiple persons with the same surname may be distinguished as ``Mou A/Mou B''. Already de-identified names should not be over-rewritten.\\
2. Addresses, including registered residence, current residence, home address, delivery address, private residence, house number, building, unit, room number, and village group. Province, city, county/district, township/street, and other coarse locations may be kept; private precise details should be masked.\\
3. Full or partially masked ID numbers, phone numbers, birth dates, bank-card/account numbers, Alipay/WeChat/QQ/email, express/logistics numbers, full license plate numbers, and identifiable private company/shop/school names.\\
{}\textbf{[Do not change]}\\
Do not change case number, metadata, source URL, decision, relevant articles, charges, amounts, quantities, incident dates, procedural dates, law articles, evidence types, litigation procedure, or disposition. Do not change ``suspect'' into ``defendant'' or ``non-prosecuted person''. Do not summarize, delete, or rewrite facts for de-identification when local replacement is possible. Do not de-identify legal organs, forensic centers, public security bureaus, procuratorates, courts, administrative divisions, or province/city/county names.\\
{}\textbf{[Consistency requirements]}\\
Use the same de-identified name for the same natural person across the four fields. Do not merge different natural persons into one appellation. If the input has already been partially de-identified, only complete the remaining identifiable parts.\\
{}\textbf{[Final output requirements]}\\
The final JSON must contain only the following four keys, with exactly matching key names and types: \{``person\_info'':``...'', ``procedure'':``...'', ``fact'':``...'', ``raw\_reasoning\_and\_decision'':``...''\}. If you output a thinking process, the JSON must appear last and nothing may follow it.\\
{}\textbf{[JSON to be de-identified]} \{record\_json\}
\end{tcolorbox}
\caption{User prompt template for sensitive-information de-identification.}
\label{fig:app_deid_user_prompt}
\end{figure*}

\section{Model-Level Correlations Between LJP and PDP Class Metrics}
\label{app:ljp_pdp_correlation}

\begin{figure*}[t]
    \centering
    \includegraphics[width=\textwidth]{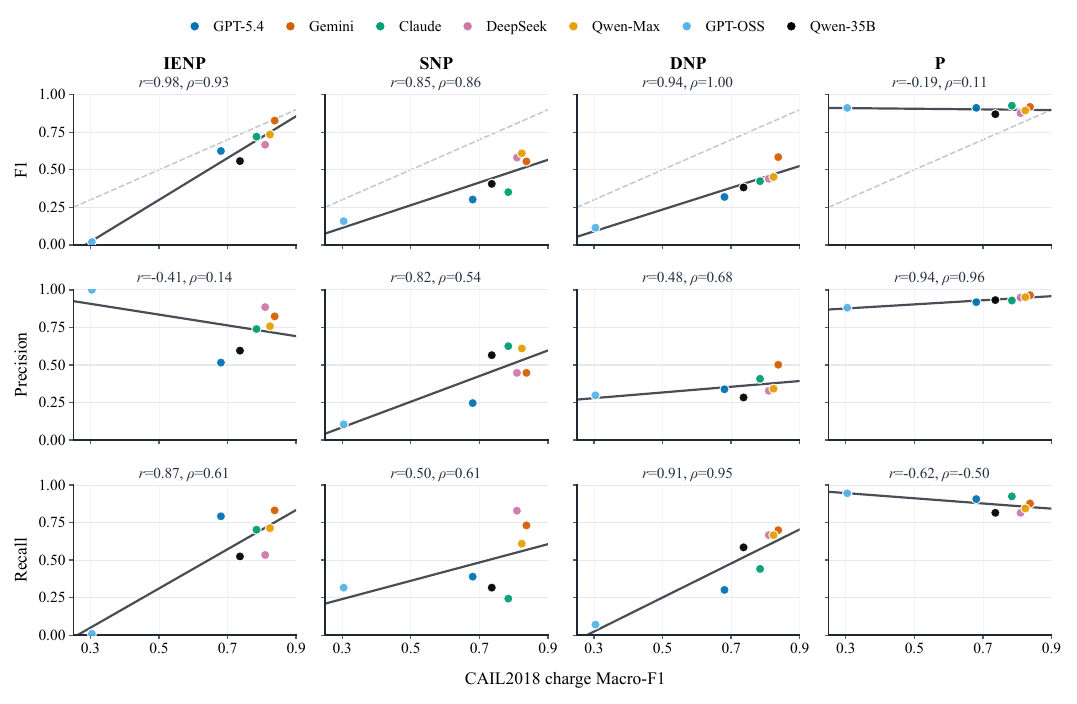}
    \caption{Model-level correlations between CAIL2018 charge-prediction Macro-F1 and class-level F1, precision, and recall on PDP Level 2. Each point denotes a model; colors distinguish models. Solid lines are least-squares fits, and the dashed line in the first row is the $y=x$ reference. $r$ and $\rho$ denote Pearson and Spearman correlations.}
    \label{fig:app_ljp_pdp_correlation}
\end{figure*}

CAIL2018 charge prediction, our LJP reference, is strongly and positively correlated with the three PDP Level 2 NP subclasses at the model level. Thus, conventional LJP scores still capture part of a model's general legal-text discrimination ability. Yet correlation is not coverage: most models fall below the $y=x$ reference line in the IENP, SNP, and DNP panels, and the best SNP/DNP F1 scores reach only 0.6098/0.5843. PDP therefore remains difficult even for strong LJP models.

Precision and recall reveal different error structures across classes. P is especially revealing: its precision is strongly correlated with CAIL ($r=0.94$, $\rho=0.96$), but its recall is negatively correlated ($r=-0.62$, $\rho=-0.50$), leaving P-class F1 almost uncorrelated with CAIL Macro-F1 ($r=-0.19$, $\rho=0.11$). The main source of model differentiation in PDP is therefore not P, which conventional LJP already covers relatively well, but the evidence-evaluation, legal-subsumption, and value-discretion boundaries represented by IENP, SNP, and DNP.

\section{Per-Class Failure Analysis on PDP}
\label{app:failure_analysis}

To characterize how SOTA LLMs go wrong on each PDP class, we inspect gold-vs-prediction confusion entries from Gemini-3.1-Pro---the strongest model on PDP-Bench.test---and report a representative case for each of the three non-prosecution classes.

\paragraph{IENP example: evidence does not exclude reasonable doubt.}
A county livestock-bureau official helped his nephew obtain 100{,}000 yuan in livestock-subsidy funding by certifying a sheep farm that did not meet the regulation's conditions. Investigators charged joint embezzlement under Criminal Law (CL) Articles~25/382. After two rounds of supplementary investigation, the procuratorate held that the available evidence could not prove the official's subjective intent to unlawfully appropriate the funds beyond reasonable doubt: the suspect denied intent, never personally possessed or controlled the disbursed funds, and the only direct evidence on a related bribery allegation was a one-sided statement from the alleged briber---an isolated-witness situation. The procuratorate returned an IENP under Criminal Procedure Law (CPL) Article~171(4). The model recognized the same evidence chain but treated the circumstantial pattern (organising the application and leading the verification) as conclusive of intent, recommending prosecution.

\paragraph{SNP example: misclassification of actor role under the corruption statutes.}
A village officer fabricated 47{,}550 yuan in coal-bed-gas land-use compensation records over multiple years. The model classified the suspect as a state functionary under CL Articles~382/93 (corruption, with a 30{,}000-yuan threshold) and recommended prosecution. The procuratorate held that helping distribute coal-bed-gas compensation does not fall within the legislative interpretation of Article~93 (which lists ``assisting government in land-acquisition compensation management''), so the conduct is governed by CL Article~271 (embezzlement by company personnel, with a 60{,}000-yuan threshold). Because the amount falls below this threshold, the conduct does not constitute a crime, and the procuratorate returned an SNP under CPL Articles~15(6)/173(1).

\paragraph{DNP example: leniency factors recognized but not weighed as punishment-waiving.}
A driver caused a fatal traffic accident, voluntarily surrendered to police, compensated the victim's family, and obtained their forgiveness. The model recognized the crime (traffic casualty, CL Article~133) and the self-surrender (CL Article~67(1)) but treated the leniency factors as sentence-mitigating signals only and recommended prosecution. The procuratorate combined the same factors with CL Article~37 (exemption from punishment for minor circumstances) and returned a DNP under CPL Article~173(2).

Across the three cases, the model's reasoning stops one step short of the procuratorate's: it identifies the relevant evidence/articles/factors but does not perform the additional checks of evidence-chain closure, fine-grained statute selection, or punishment waiver under Article~37. These steps correspond precisely to PDP's evidence evaluation, legal subsumption, and value-based discretion.

\section{Reasoning-Budget Levels}
\label{sec:app_budget_levels}

In Section~\ref{sec:test_time_scaling}, the five models are intervened through their natively supported OpenRouter reasoning-budget interfaces. Table~\ref{tab:app_budget_levels} reports whether each model uses an effort-level interface or a token-budget interface, together with its valid levels.

\begin{table}[!htbp]
\centering
\footnotesize
\setlength{\tabcolsep}{3pt}
\resizebox{\columnwidth}{!}{%
\begin{tabular}{@{}lll@{}}
\toprule
\textbf{Model} & \textbf{Budget} & \textbf{Levels} \\
\midrule
GPT-5.4 & effort & none / low / medium / high / xhigh \\
Gemini-3.1-Pro & effort & low / medium / high \\
DeepSeek-V4-Pro & effort & none / high / xhigh \\
Claude-Opus-4.6 & max\_tokens & 0 / 1024 / 2048 / 4096 \\
Qwen3.6-Max & max\_tokens & 0 / 512 / 1024 / 2048 / 4096 \\
\bottomrule
\end{tabular}%
}
\caption{Reasoning-budget levels used in test-time scaling.}
\label{tab:app_budget_levels}
\end{table}

In OpenRouter, effort and max\_tokens are mutually exclusive, and supported levels differ across models. GPT-5.4 does not support some combinations below low/medium and none; Gemini-3.1-Pro does not accept effort=none; and DeepSeek-V4-Pro does not accept effort=low/medium. We therefore restrict the intervention levels to each model's valid native set.

Section~\ref{sec:test_time_scaling} uses \emph{test\_rq2}, a 100-sample subset of PDP-Bench formed by uniformly sampling 25 instances per decision class. Each sample is inferred 8 times independently, yielding 800 inferences per budget level.

\section{Full Reasoning-Budget Results}
\label{sec:app_budget_full}

Tables~\ref{tab:app_budget_l1} and \ref{tab:app_budget_l2} report the full overall and class-level F1-score results for the five SOTA models under different reasoning budgets. Values are means over 8 runs; the number after $\pm$ is the standard error computed from run-level metrics (SE, $n=8$, $df=7$).

\begin{table*}[t]
\centering
\scriptsize
\resizebox{\textwidth}{!}{%
\begin{tabular}{@{}llcccc@{}}
\toprule
\textbf{Model} & \textbf{Budget} & \textbf{Macro-F1} & \textbf{Micro-F1} & \textbf{NP} & \textbf{P} \\
\midrule
\multirow{5}{*}{GPT-5.4} & none & 0.7686 $\pm$ 0.0023 & 0.7962 $\pm$ 0.0023 & 0.8517 $\pm$ 0.0023 & 0.6854 $\pm$ 0.0035 \\
 & low & 0.7284 $\pm$ 0.0049 & 0.7522 $\pm$ 0.0055 & 0.8112 $\pm$ 0.0054 & 0.6457 $\pm$ 0.0054 \\
 & medium & 0.7143 $\pm$ 0.0066 & 0.7394 $\pm$ 0.0058 & 0.8031 $\pm$ 0.0040 & 0.6255 $\pm$ 0.0106 \\
 & high & 0.7242 $\pm$ 0.0049 & 0.7479 $\pm$ 0.0052 & 0.8065 $\pm$ 0.0053 & 0.6419 $\pm$ 0.0061 \\
 & xhigh & 0.4546 $\pm$ 0.0174 & 0.4915 $\pm$ 0.0143 & 0.5519 $\pm$ 0.0169 & 0.3574 $\pm$ 0.0316 \\
\midrule
\multirow{3}{*}{Gemini-3.1-Pro} & low & 0.8324 $\pm$ 0.0055 & 0.8640 $\pm$ 0.0044 & 0.9060 $\pm$ 0.0030 & 0.7588 $\pm$ 0.0084 \\
 & medium & 0.8517 $\pm$ 0.0070 & 0.8787 $\pm$ 0.0056 & 0.9143 $\pm$ 0.0049 & 0.7891 $\pm$ 0.0113 \\
 & high & 0.8451 $\pm$ 0.0068 & 0.8701 $\pm$ 0.0075 & 0.9019 $\pm$ 0.0085 & 0.7883 $\pm$ 0.0073 \\
\midrule
\multirow{4}{*}{Claude-Opus-4.6} & 0 & 0.8118 $\pm$ 0.0053 & 0.8373 $\pm$ 0.0050 & 0.8817 $\pm$ 0.0040 & 0.7418 $\pm$ 0.0067 \\
 & 1024 & 0.8209 $\pm$ 0.0042 & 0.8462 $\pm$ 0.0037 & 0.8883 $\pm$ 0.0028 & 0.7536 $\pm$ 0.0056 \\
 & 2048 & 0.8043 $\pm$ 0.0029 & 0.8312 $\pm$ 0.0030 & 0.8769 $\pm$ 0.0024 & 0.7317 $\pm$ 0.0034 \\
 & 4096 & 0.8104 $\pm$ 0.0037 & 0.8355 $\pm$ 0.0035 & 0.8791 $\pm$ 0.0031 & 0.7416 $\pm$ 0.0047 \\
\midrule
\multirow{3}{*}{DeepSeek-V4-Pro} & none & 0.7840 $\pm$ 0.0152 & 0.8280 $\pm$ 0.0129 & 0.8839 $\pm$ 0.0091 & 0.6840 $\pm$ 0.0218 \\
 & high & 0.7529 $\pm$ 0.0082 & 0.7893 $\pm$ 0.0075 & 0.8475 $\pm$ 0.0061 & 0.6584 $\pm$ 0.0115 \\
 & xhigh & 0.7735 $\pm$ 0.0080 & 0.8091 $\pm$ 0.0080 & 0.8628 $\pm$ 0.0072 & 0.6842 $\pm$ 0.0101 \\
\midrule
\multirow{5}{*}{Qwen3.6-Max} & 0 & 0.7468 $\pm$ 0.0152 & 0.7730 $\pm$ 0.0153 & 0.8275 $\pm$ 0.0130 & 0.6660 $\pm$ 0.0178 \\
 & 512 & 0.8041 $\pm$ 0.0151 & 0.8384 $\pm$ 0.0135 & 0.8872 $\pm$ 0.0102 & 0.7211 $\pm$ 0.0201 \\
 & 1024 & 0.7851 $\pm$ 0.0053 & 0.8173 $\pm$ 0.0059 & 0.8678 $\pm$ 0.0056 & 0.7024 $\pm$ 0.0061 \\
 & 2048 & 0.7939 $\pm$ 0.0074 & 0.8223 $\pm$ 0.0071 & 0.8703 $\pm$ 0.0059 & 0.7175 $\pm$ 0.0091 \\
 & 4096 & 0.7915 $\pm$ 0.0096 & 0.8225 $\pm$ 0.0094 & 0.8718 $\pm$ 0.0075 & 0.7112 $\pm$ 0.0118 \\
\bottomrule
\end{tabular}%
}
\caption{L1 reasoning-budget results across models.}
\label{tab:app_budget_l1}
\end{table*}

\begin{table*}[t]
\centering
\scriptsize
\resizebox{\textwidth}{!}{%
\begin{tabular}{@{}llccccc@{}}
\toprule
\textbf{Model} & \textbf{Budget} & \textbf{Macro-F1} & \textbf{Micro-F1} & \textbf{IENP} & \textbf{SNP} & \textbf{DNP} \\
\midrule
\multirow{5}{*}{GPT-5.4} & none & 0.5912 $\pm$ 0.0060 & 0.6101 $\pm$ 0.0046 & 0.6586 $\pm$ 0.0106 & 0.5610 $\pm$ 0.0195 & 0.4597 $\pm$ 0.0145 \\
 & low & 0.6228 $\pm$ 0.0072 & 0.6377 $\pm$ 0.0055 & 0.7345 $\pm$ 0.0139 & 0.7347 $\pm$ 0.0132 & 0.3762 $\pm$ 0.0179 \\
 & medium & 0.6411 $\pm$ 0.0047 & 0.6510 $\pm$ 0.0054 & 0.7573 $\pm$ 0.0114 & 0.8023 $\pm$ 0.0151 & 0.3793 $\pm$ 0.0115 \\
 & high & 0.6388 $\pm$ 0.0104 & 0.6493 $\pm$ 0.0093 & 0.7548 $\pm$ 0.0101 & 0.7696 $\pm$ 0.0218 & 0.3888 $\pm$ 0.0139 \\
 & xhigh & 0.4238 $\pm$ 0.0131 & 0.4293 $\pm$ 0.0142 & 0.4834 $\pm$ 0.0304 & 0.6197 $\pm$ 0.0267 & 0.2346 $\pm$ 0.0133 \\
\midrule
\multirow{3}{*}{Gemini-3.1-Pro} & low & 0.7849 $\pm$ 0.0043 & 0.7812 $\pm$ 0.0048 & 0.8909 $\pm$ 0.0070 & 0.8386 $\pm$ 0.0066 & 0.6513 $\pm$ 0.0077 \\
 & medium & 0.8073 $\pm$ 0.0058 & 0.8055 $\pm$ 0.0061 & 0.8975 $\pm$ 0.0053 & 0.8473 $\pm$ 0.0074 & 0.6955 $\pm$ 0.0093 \\
 & high & 0.8028 $\pm$ 0.0072 & 0.8011 $\pm$ 0.0070 & 0.8834 $\pm$ 0.0133 & 0.8532 $\pm$ 0.0119 & 0.6865 $\pm$ 0.0132 \\
\midrule
\multirow{4}{*}{Claude-Opus-4.6} & 0 & 0.5940 $\pm$ 0.0056 & 0.6145 $\pm$ 0.0050 & 0.7230 $\pm$ 0.0085 & 0.4467 $\pm$ 0.0152 & 0.4644 $\pm$ 0.0043 \\
 & 1024 & 0.6673 $\pm$ 0.0049 & 0.6787 $\pm$ 0.0044 & 0.7584 $\pm$ 0.0036 & 0.6175 $\pm$ 0.0096 & 0.5396 $\pm$ 0.0085 \\
 & 2048 & 0.6609 $\pm$ 0.0045 & 0.6700 $\pm$ 0.0038 & 0.7567 $\pm$ 0.0055 & 0.6353 $\pm$ 0.0123 & 0.5200 $\pm$ 0.0064 \\
 & 4096 & 0.6564 $\pm$ 0.0064 & 0.6679 $\pm$ 0.0057 & 0.7584 $\pm$ 0.0085 & 0.6119 $\pm$ 0.0168 & 0.5135 $\pm$ 0.0058 \\
\midrule
\multirow{3}{*}{DeepSeek-V4-Pro} & none & 0.7092 $\pm$ 0.0143 & 0.7054 $\pm$ 0.0146 & 0.7373 $\pm$ 0.0212 & 0.7577 $\pm$ 0.0198 & 0.6577 $\pm$ 0.0178 \\
 & high & 0.7004 $\pm$ 0.0086 & 0.6991 $\pm$ 0.0077 & 0.6644 $\pm$ 0.0219 & 0.8315 $\pm$ 0.0135 & 0.6473 $\pm$ 0.0088 \\
 & xhigh & 0.7159 $\pm$ 0.0110 & 0.7140 $\pm$ 0.0108 & 0.6865 $\pm$ 0.0149 & 0.8308 $\pm$ 0.0121 & 0.6621 $\pm$ 0.0223 \\
\midrule
\multirow{5}{*}{Qwen3.6-Max} & 0 & 0.6163 $\pm$ 0.0156 & 0.6217 $\pm$ 0.0152 & 0.6837 $\pm$ 0.0237 & 0.5943 $\pm$ 0.0268 & 0.5211 $\pm$ 0.0203 \\
 & 512 & 0.6753 $\pm$ 0.0140 & 0.6767 $\pm$ 0.0144 & 0.7452 $\pm$ 0.0129 & 0.6350 $\pm$ 0.0141 & 0.6001 $\pm$ 0.0229 \\
 & 1024 & 0.6980 $\pm$ 0.0098 & 0.6956 $\pm$ 0.0092 & 0.7477 $\pm$ 0.0154 & 0.7522 $\pm$ 0.0200 & 0.5896 $\pm$ 0.0140 \\
 & 2048 & 0.7181 $\pm$ 0.0064 & 0.7159 $\pm$ 0.0068 & 0.7771 $\pm$ 0.0101 & 0.8074 $\pm$ 0.0122 & 0.5704 $\pm$ 0.0095 \\
 & 4096 & 0.7244 $\pm$ 0.0104 & 0.7200 $\pm$ 0.0105 & 0.7940 $\pm$ 0.0143 & 0.8177 $\pm$ 0.0162 & 0.5746 $\pm$ 0.0127 \\
\bottomrule
\end{tabular}%
}
\caption{L2 reasoning-budget results across models.}
\label{tab:app_budget_l2}
\end{table*}

The complete estimates reinforce the pattern in Section~\ref{sec:test_time_scaling}. Reasoning budget is not a monotonic control knob: moderate budgets improve L2 performance for Gemini-3.1-Pro, Claude-Opus-4.6, and Qwen3.6-Max, but GPT-5.4 collapses under xhigh and DeepSeek-V4-Pro remains comparatively flat. The class-level rows also show that gains on \textsc{IENP} or \textsc{SNP} do not reliably transfer to \textsc{DNP}, which is the most persistent boundary under test-time scaling.

\section{Full Legal-Domain Specialization Results}
\label{sec:app_legal_full}

Tables~\ref{tab:app_legal_se} and \ref{tab:app_legal_se_l2} report the full overall and class-level F1 scores for the three base--legal model pairs in Section~\ref{sec:legal_specialization}. Values are means over 8 runs; the number after $\pm$ is the standard error computed from run-level metrics (SE, $n=8$, $df=7$).

\begin{table*}[t]
\centering
\scriptsize
\resizebox{\textwidth}{!}{%
\begin{tabular}{@{}lllcccc@{}}
\toprule
\textbf{Training Paradigm} & \textbf{Model} & \textbf{Type} & \textbf{Macro-F1} & \textbf{Micro-F1} & \textbf{NP} & \textbf{P} \\
\midrule
\multirow{2}{*}{Full SFT} & Qwen2.5-7B-Instruct & Base & 0.6274 $\pm$ 0.0053 & 0.7923 $\pm$ 0.0072 & 0.3825 $\pm$ 0.0066 & 0.8723 $\pm$ 0.0050 \\
 & LawLLM-7B & Legal & 0.3947 $\pm$ 0.0050 & 0.6062 $\pm$ 0.0076 & 0.1271 $\pm$ 0.0057 & 0.6622 $\pm$ 0.0081 \\
\midrule
\multirow{2}{*}{R1-Distill SFT+GRPO} & Qwen2.5-14B-Instruct & Base & 0.5779 $\pm$ 0.0067 & 0.6657 $\pm$ 0.0106 & 0.3937 $\pm$ 0.0067 & 0.7621 $\pm$ 0.0101 \\
 & LegalDelta & Legal & 0.5583 $\pm$ 0.0105 & 0.6559 $\pm$ 0.0119 & 0.3528 $\pm$ 0.0131 & 0.7638 $\pm$ 0.0097 \\
\midrule
\multirow{3}{*}{Mid-training+LEAD-SFT+RL} & Qwen3-8B-Base & Base & 0.5314 $\pm$ 0.0047 & 0.7405 $\pm$ 0.0058 & 0.2386 $\pm$ 0.0080 & 0.8242 $\pm$ 0.0043 \\
 & Qwen3-8B & Instruct & 0.6160 $\pm$ 0.0047 & 0.8657 $\pm$ 0.0010 & 0.3065 $\pm$ 0.0091 & 0.9255 $\pm$ 0.0005 \\
 & LegalOne-R1-8B & Legal & 0.6123 $\pm$ 0.0018 & 0.7277 $\pm$ 0.0030 & 0.4169 $\pm$ 0.0022 & 0.8077 $\pm$ 0.0039 \\
\bottomrule
\end{tabular}%
}
\caption{L1 overall results with standard errors for legal-domain specialization.}
\label{tab:app_legal_se}
\end{table*}

\begin{table*}[t]
\centering
\scriptsize
\resizebox{\textwidth}{!}{%
\begin{tabular}{@{}lllccccc@{}}
\toprule
\textbf{Training Paradigm} & \textbf{Model} & \textbf{Type} & \textbf{Macro-F1} & \textbf{Micro-F1} & \textbf{IENP} & \textbf{SNP} & \textbf{DNP} \\
\midrule
\multirow{2}{*}{Full SFT} & Qwen2.5-7B-Instruct & Base & 0.4161 $\pm$ 0.0041 & 0.7778 $\pm$ 0.0072 & 0.2923 $\pm$ 0.0132 & 0.1972 $\pm$ 0.0106 & 0.3026 $\pm$ 0.0066 \\
 & LawLLM-7B & Legal & 0.2120 $\pm$ 0.0037 & 0.6009 $\pm$ 0.0074 & 0.0399 $\pm$ 0.0060 & 0.0589 $\pm$ 0.0105 & 0.0869 $\pm$ 0.0052 \\
\midrule
\multirow{2}{*}{R1-Distill SFT+GRPO} & Qwen2.5-14B-Instruct & Base & 0.4543 $\pm$ 0.0063 & 0.6514 $\pm$ 0.0106 & 0.4702 $\pm$ 0.0114 & 0.2548 $\pm$ 0.0262 & 0.3301 $\pm$ 0.0080 \\
 & LegalDelta & Legal & 0.4631 $\pm$ 0.0105 & 0.6410 $\pm$ 0.0120 & 0.3957 $\pm$ 0.0107 & 0.4060 $\pm$ 0.0276 & 0.2867 $\pm$ 0.0137 \\
\midrule
\multirow{3}{*}{Mid-training+LEAD-SFT+RL} & Qwen3-8B-Base & Base & 0.2967 $\pm$ 0.0081 & 0.7312 $\pm$ 0.0063 & 0.0594 $\pm$ 0.0102 & 0.1043 $\pm$ 0.0192 & 0.1991 $\pm$ 0.0084 \\
 & Qwen3-8B & Instruct & 0.4546 $\pm$ 0.0051 & 0.8601 $\pm$ 0.0009 & 0.3440 $\pm$ 0.0126 & 0.3273 $\pm$ 0.0133 & 0.2216 $\pm$ 0.0114 \\
 & LegalOne-R1-8B & Legal & 0.5391 $\pm$ 0.0053 & 0.7212 $\pm$ 0.0031 & 0.4202 $\pm$ 0.0145 & 0.5534 $\pm$ 0.0178 & 0.3749 $\pm$ 0.0023 \\
\bottomrule
\end{tabular}%
}
\caption{L2 overall results with standard errors for legal-domain specialization.}
\label{tab:app_legal_se_l2}
\end{table*}

These full estimates make the source of the legal-specialization result more explicit. Generic legal SFT can substantially hurt PDP, and R1-Distill SFT+GRPO does not yield a stable legal-model advantage. The strongest L2 improvement appears only under the Mid-training+LEAD-SFT+RL recipe, suggesting that PDP benefits from the training pipeline rather than from legal-domain labeling alone.

\section{Full Prompt-Augmentation Results}
\label{sec:app_prompt_full}

Tables~\ref{tab:app_prompt_l1} and \ref{tab:app_prompt_l2} report the full overall and class-level F1-score results for the five SOTA models under the Baseline, +Definitions, and +One-shot prompt conditions in Section~\ref{sec:prompt_aug}. Values are means over 8 runs; the number after $\pm$ is the standard error computed from run-level metrics (SE, $n=8$, $df=7$).

\subsection{Prompt Settings}
\label{sec:app_prompts}

All PDP experiments use the same input fields -- suspect information, procedural information, and factual information -- and require the model to output applicable law articles, a review analysis, and one final decision from P, \textsc{DNP}, \textsc{SNP}, and \textsc{IENP}. Unless explicitly varied, the prompt is the Baseline prompt. Thus, Section~\ref{sec:rq1}, Section~\ref{sec:test_time_scaling}, Section~\ref{sec:legal_specialization}, and Section~\ref{sec:rq3} change the model, reasoning budget, model specialization, or training split while keeping the base task instruction fixed. Section~\ref{sec:prompt_aug} is the only experiment that varies the prompt itself. The following boxes give English renderings of the prompts used in the implementation.

\paragraph{Baseline.}
The baseline condition contains only the role instruction, the four admissible decisions, the required output schema, and the three case fields. It provides no additional legal definitions or demonstrations, so it serves as the common task instruction against which the two prompt-augmentation variants are compared.

\begin{figure*}[t]
\begin{tcolorbox}[enhanced jigsaw, title=Baseline Prompt, colback=white, colframe=black!35, boxrule=0.4pt, arc=1mm, left=1mm, right=1mm, top=1mm, bottom=1mm]
\footnotesize
\textbf{System Prompt}\\
You are an intelligent assistant for prosecutors, familiar with the Criminal Law, the Criminal Procedure Law, and the Rules of Criminal Procedure for People's Procuratorates.\\
Given the suspect information, procedural information, and factual information, identify applicable law articles, analyze the case, and recommend a prosecution decision.\\
\textbf{Rules.}\\
\hspace*{1em} -- The decision must be one of four options: prosecution, discretionary non-prosecution, statutory non-prosecution, insufficient-evidence non-prosecution.\\
\hspace*{1em} -- Multiple law articles must be separated by the Chinese enumeration comma (U+3001).\\
The review analysis must cite the applicable law articles. Output strictly in the following format:\\
\hspace*{1em} [Applicable law articles] (list cited articles in Chinese numerals; sub-categories with no citation may be omitted)\\
\hspace*{2em} Criminal Law: Article XXX, Article XXX(XXX)\\
\hspace*{2em} Criminal Procedure Law: Article XXX, Article XXX(XXX)\\
\hspace*{2em} Rules of Criminal Procedure: Article XXX, Article XXX\\
\hspace*{1em} [Review analysis] (cite the applicable law articles and analyze the case facts)\\
\hspace*{1em} [Final conclusion] Decision: prosecution / discretionary non-prosecution / statutory non-prosecution / insufficient-evidence non-prosecution (choose one)\\

- - -\\
\textbf{User Prompt}\\
Please review the following case and give a prosecution decision.\\
Suspect information:\\
\hspace*{1em} \{person\_info\}\\
Procedural information:\\
\hspace*{1em} \{procedure\}\\
Factual information:\\
\hspace*{1em} \{fact\}
\end{tcolorbox}
\caption{Baseline task prompt used when no prompt augmentation is applied.}
\label{fig:app_prompt_baseline}
\end{figure*}

\paragraph{+Definitions.}
This condition keeps the same case-input format as Baseline, but appends the following definition block to the system instruction.

\begin{figure*}[t]
\begin{tcolorbox}[enhanced jigsaw, title=Additional Definition Block, colback=white, colframe=black!35, boxrule=0.4pt, arc=1mm, left=1mm, right=1mm, top=1mm, bottom=1mm]
\footnotesize
\textbf{Decision Definitions}\\
1. Prosecution\\
\hspace*{1em} Legal basis: Article 176 of the Criminal Procedure Law.\\
\hspace*{1em} Applicability: the criminal facts have been ascertained, the evidence is reliable and sufficient, and criminal liability should be pursued.\\

2. Statutory non-prosecution\\
\hspace*{1em} Legal basis: Article 177(1) and Article 16 of the Criminal Procedure Law.\\
\hspace*{1em} Applicability: there is no criminal fact, or criminal liability is legally barred, including obviously minor circumstances not deemed criminal, expiration of the limitation period, special amnesty, withdrawal or absence of complaint for complaint-only offenses, death of the suspect or defendant, or other statutory grounds for exemption from liability.\\

3. Insufficient-evidence non-prosecution\\
\hspace*{1em} Legal basis: Article 175(4) of the Criminal Procedure Law.\\
\hspace*{1em} Applicability: after supplementary investigation, the evidence remains insufficient and the conditions for prosecution are not met. This includes doubtful evidence, missing proof for offense elements, unresolved contradictions among evidence, or alternative explanations that cannot be excluded beyond reasonable doubt.\\

4. Discretionary non-prosecution\\
\hspace*{1em} Legal basis: Article 177(2) of the Criminal Procedure Law and Article 370 of the Rules of Criminal Procedure for People's Procuratorates.\\
\hspace*{1em} Applicability: the facts are clear, the evidence is reliable and sufficient, and the conduct constitutes a crime, but the circumstances are minor and punishment is unnecessary or may be exempted under the Criminal Law. Judgement should distinguish this category from the other two non-prosecution categories: (i) a criminal fact exists and no Article-16 ground for exemption applies (so not statutory non-prosecution); (ii) the evidence is sufficient and reasonable doubt has been excluded (so not insufficient-evidence non-prosecution); (iii) leniency factors -- voluntary surrender, truthful confession, guilty plea and acceptance of punishment, active return of illegal gains, compensation with victim forgiveness, first-time or occasional offence, minor harmful consequences -- are typically considered together, but these factors alone cannot substitute for ``minor circumstances'' and ``no need for punishment or punishment may be lawfully exempted''.
\end{tcolorbox}
\caption{Additional legal-definition block used in the +Definitions prompt condition.}
\label{fig:app_prompt_definitions}
\end{figure*}

\paragraph{+One-shot.}
This condition keeps the Baseline system instruction. The one-shot demonstration is delivered as a two-turn user/assistant exchange before the actual test case, so the chat history contains four messages in total (system, example user, example assistant, real user).

\begin{figure*}[t]
\begin{tcolorbox}[enhanced jigsaw, title=One-Shot Demonstration (delivered as four chat messages), colback=white, colframe=black!35, boxrule=0.4pt, arc=1mm, left=1mm, right=1mm, top=1mm, bottom=1mm]
\footnotesize
\textbf{Message 1 -- system}\\
\hspace*{1em} (identical to the Baseline system prompt above)\\

- - -\\
\textbf{Message 2 -- user (example input)}\\
Please review the following case and give a prosecution decision.\\
Suspect information:\\
\hspace*{1em} The suspect, an adult male with an elementary-school education, resides in a county-level locality.\\
Procedural information:\\
\hspace*{1em} The suspect was criminally detained for suspected dangerous driving on August 21, 2023 by the local public security bureau, and released on bail pending trial on August 24, 2023 by the local people's procuratorate.\\
Factual information:\\
\hspace*{1em} At 18:55 on August 17, 2023, the suspect drove an unlicensed two-wheeled motorcycle on a scenic highway and was stopped by traffic police. A forensic test by a public-security traffic forensic center showed a blood alcohol concentration of 107.84 mg/100 ml.\\

- - -\\
\textbf{Message 3 -- assistant (example output)}\\
{}[Applicable law articles]\\
\hspace*{1em} Criminal Law: Article 133-1, Article 37.\\
\hspace*{1em} Criminal Procedure Law: Article 177(2).\\
{}[Review analysis]\\
\hspace*{1em} The suspect committed the act covered by Article 133-1 of the Criminal Law. However, the circumstances are minor, no serious social harm was caused, and the suspect admitted guilt and accepted punishment. Under Article 37 of the Criminal Law, criminal punishment is unnecessary. We recommend discretionary non-prosecution under Article 177(2) of the Criminal Procedure Law.\\
{}[Final conclusion]\\
\hspace*{1em} Decision: discretionary non-prosecution.\\

- - -\\
\textbf{Message 4 -- user (test case)}\\
\hspace*{1em} (the actual case to be reviewed, in the same user-prompt format as Message 2)
\end{tcolorbox}
\caption{One-shot demonstration used in the +One-shot prompt condition.}
\label{fig:app_prompt_oneshot}
\end{figure*}

\begin{table*}[t]
\centering
\scriptsize
\resizebox{\textwidth}{!}{%
\begin{tabular}{@{}llcccc@{}}
\toprule
\textbf{Model} & \textbf{Prompt} & \textbf{Macro-F1} & \textbf{Micro-F1} & \textbf{NP} & \textbf{P} \\
\midrule
\multirow{3}{*}{GPT-5.4} & Baseline & 0.7634 $\pm$ 0.0073 & 0.7892 $\pm$ 0.0064 & 0.8429 $\pm$ 0.0050 & 0.6838 $\pm$ 0.0107 \\
 & +Definitions & 0.7612 $\pm$ 0.0075 & 0.7900 $\pm$ 0.0072 & 0.8466 $\pm$ 0.0059 & 0.6758 $\pm$ 0.0094 \\
 & +One-shot & 0.7724 $\pm$ 0.0061 & 0.7972 $\pm$ 0.0061 & 0.8497 $\pm$ 0.0054 & 0.6952 $\pm$ 0.0073 \\
\midrule
\multirow{3}{*}{Gemini-3.1-Pro} & Baseline & 0.8146 $\pm$ 0.0089 & 0.8426 $\pm$ 0.0074 & 0.8783 $\pm$ 0.0058 & 0.7509 $\pm$ 0.0130 \\
 & +Definitions & 0.8220 $\pm$ 0.0064 & 0.8572 $\pm$ 0.0062 & 0.8990 $\pm$ 0.0066 & 0.7449 $\pm$ 0.0092 \\
 & +One-shot & 0.8642 $\pm$ 0.0043 & 0.8894 $\pm$ 0.0045 & 0.9201 $\pm$ 0.0049 & 0.8082 $\pm$ 0.0053 \\
\midrule
\multirow{3}{*}{Claude-Opus-4.6} & Baseline & 0.8063 $\pm$ 0.0026 & 0.8325 $\pm$ 0.0025 & 0.8775 $\pm$ 0.0020 & 0.7352 $\pm$ 0.0035 \\
 & +Definitions & 0.8035 $\pm$ 0.0041 & 0.8300 $\pm$ 0.0038 & 0.8757 $\pm$ 0.0029 & 0.7313 $\pm$ 0.0052 \\
 & +One-shot & 0.7891 $\pm$ 0.0066 & 0.8170 $\pm$ 0.0059 & 0.8671 $\pm$ 0.0041 & 0.7111 $\pm$ 0.0094 \\
\midrule
\multirow{3}{*}{DeepSeek-V4-Pro} & Baseline & 0.7567 $\pm$ 0.0072 & 0.7914 $\pm$ 0.0063 & 0.8494 $\pm$ 0.0046 & 0.6639 $\pm$ 0.0102 \\
 & +Definitions & 0.7127 $\pm$ 0.0116 & 0.7430 $\pm$ 0.0119 & 0.8048 $\pm$ 0.0109 & 0.6205 $\pm$ 0.0136 \\
 & +One-shot & 0.7809 $\pm$ 0.0077 & 0.8241 $\pm$ 0.0064 & 0.8798 $\pm$ 0.0046 & 0.6819 $\pm$ 0.0112 \\
\midrule
\multirow{3}{*}{Qwen3.6-Max} & Baseline & 0.7901 $\pm$ 0.0070 & 0.8192 $\pm$ 0.0071 & 0.8678 $\pm$ 0.0061 & 0.7123 $\pm$ 0.0083 \\
 & +Definitions & 0.8059 $\pm$ 0.0083 & 0.8405 $\pm$ 0.0078 & 0.8875 $\pm$ 0.0058 & 0.7242 $\pm$ 0.0109 \\
 & +One-shot & 0.8529 $\pm$ 0.0087 & 0.8825 $\pm$ 0.0075 & 0.9188 $\pm$ 0.0054 & 0.7871 $\pm$ 0.0120 \\
\bottomrule
\end{tabular}%
}
\caption{L1 prompt-augmentation results across models.}
\label{tab:app_prompt_l1}
\end{table*}

\begin{table*}[t]
\centering
\scriptsize
\resizebox{\textwidth}{!}{%
\begin{tabular}{@{}llccccc@{}}
\toprule
\textbf{Model} & \textbf{Prompt} & \textbf{Macro-F1} & \textbf{Micro-F1} & \textbf{IENP} & \textbf{SNP} & \textbf{DNP} \\
\midrule
\multirow{3}{*}{GPT-5.4} & Baseline & 0.6006 $\pm$ 0.0067 & 0.6180 $\pm$ 0.0067 & 0.6853 $\pm$ 0.0076 & 0.5699 $\pm$ 0.0105 & 0.4635 $\pm$ 0.0122 \\
 & +Definitions & 0.6220 $\pm$ 0.0055 & 0.6453 $\pm$ 0.0049 & 0.7665 $\pm$ 0.0099 & 0.6317 $\pm$ 0.0088 & 0.4142 $\pm$ 0.0200 \\
 & +One-shot & 0.5919 $\pm$ 0.0077 & 0.6177 $\pm$ 0.0072 & 0.7178 $\pm$ 0.0115 & 0.4973 $\pm$ 0.0142 & 0.4575 $\pm$ 0.0131 \\
\midrule
\multirow{3}{*}{Gemini-3.1-Pro} & Baseline & 0.7801 $\pm$ 0.0089 & 0.7779 $\pm$ 0.0093 & 0.8572 $\pm$ 0.0166 & 0.8530 $\pm$ 0.0077 & 0.6594 $\pm$ 0.0101 \\
 & +Definitions & 0.7902 $\pm$ 0.0092 & 0.7877 $\pm$ 0.0094 & 0.8839 $\pm$ 0.0139 & 0.8433 $\pm$ 0.0136 & 0.6887 $\pm$ 0.0136 \\
 & +One-shot & 0.8089 $\pm$ 0.0069 & 0.8081 $\pm$ 0.0068 & 0.8808 $\pm$ 0.0093 & 0.8220 $\pm$ 0.0083 & 0.7246 $\pm$ 0.0123 \\
\midrule
\multirow{3}{*}{Claude-Opus-4.6} & Baseline & 0.5814 $\pm$ 0.0088 & 0.6012 $\pm$ 0.0074 & 0.7233 $\pm$ 0.0090 & 0.4396 $\pm$ 0.0225 & 0.4276 $\pm$ 0.0076 \\
 & +Definitions & 0.6617 $\pm$ 0.0034 & 0.6737 $\pm$ 0.0032 & 0.8039 $\pm$ 0.0042 & 0.6036 $\pm$ 0.0094 & 0.5078 $\pm$ 0.0056 \\
 & +One-shot & 0.6285 $\pm$ 0.0088 & 0.6403 $\pm$ 0.0081 & 0.7608 $\pm$ 0.0087 & 0.5441 $\pm$ 0.0171 & 0.4982 $\pm$ 0.0130 \\
\midrule
\multirow{3}{*}{DeepSeek-V4-Pro} & Baseline & 0.7057 $\pm$ 0.0091 & 0.7039 $\pm$ 0.0093 & 0.7006 $\pm$ 0.0152 & 0.8456 $\pm$ 0.0136 & 0.6128 $\pm$ 0.0264 \\
 & +Definitions & 0.6363 $\pm$ 0.0095 & 0.6348 $\pm$ 0.0095 & 0.6592 $\pm$ 0.0159 & 0.7402 $\pm$ 0.0189 & 0.5254 $\pm$ 0.0308 \\
 & +One-shot & 0.7199 $\pm$ 0.0043 & 0.7162 $\pm$ 0.0045 & 0.7224 $\pm$ 0.0133 & 0.7911 $\pm$ 0.0082 & 0.6842 $\pm$ 0.0091 \\
\midrule
\multirow{3}{*}{Qwen3.6-Max} & Baseline & 0.7026 $\pm$ 0.0051 & 0.7004 $\pm$ 0.0055 & 0.7631 $\pm$ 0.0091 & 0.7809 $\pm$ 0.0125 & 0.5542 $\pm$ 0.0189 \\
 & +Definitions & 0.7403 $\pm$ 0.0082 & 0.7342 $\pm$ 0.0085 & 0.8140 $\pm$ 0.0110 & 0.8074 $\pm$ 0.0078 & 0.6158 $\pm$ 0.0155 \\
 & +One-shot & 0.7691 $\pm$ 0.0099 & 0.7675 $\pm$ 0.0098 & 0.8340 $\pm$ 0.0119 & 0.7806 $\pm$ 0.0125 & 0.6749 $\pm$ 0.0164 \\
\bottomrule
\end{tabular}%
}
\caption{L2 prompt-augmentation results across models.}
\label{tab:app_prompt_l2}
\end{table*}

The full prompt results show two asymmetries. At L1, +One-shot improves four of the five models, with Claude-Opus-4.6 as the exception. At L2, both prompt variants can raise Macro-F1, but their class effects differ: +Definitions is often stronger on \textsc{IENP} and \textsc{SNP}, whereas +One-shot is more consistently helpful for \textsc{DNP}. This supports the main-text conclusion that prompt augmentation can help locally but does not provide a uniform repair.

\section{Data Splits and DAPO Training Configuration}
\label{sec:app_rq3_config}

\subsection{Class-Ratio Intervention Splits}
\label{sec:app_rq3_splits}

The seven \emph{pdp2k\_rq3} splits include a four-way balanced baseline and target-ratio intervention splits. Each split contains 2{,}000 training samples. In the intervention groups, \textsc{IENP}, \textsc{SNP}, or \textsc{DNP} is selected as the target class, and its training proportion is increased to 40\% or 55\%.

\paragraph{Data source and reproducibility.}
The \emph{pdp2k\_rq3} pool is constructed from prosecutorial decisions retrieved under our institutional license to a commercial Chinese legal database (PKULaw, ``Beida Fabao''). The pool is fully disjoint from PDP-Bench---whose samples are drawn entirely from public procuratorial websites and remain openly releasable---so the auxiliary training pool does not overlap with the evaluation set used in Sections~\ref{sec:rq1}--\ref{sec:rq3}. Due to the licensing terms of the source, the \emph{pdp2k\_rq3} samples themselves cannot be redistributed; we instead document the full construction procedure (retrieval criteria, four-class label assignment, deduplication against PDP-Bench, and the target-ratio rebalancing rule in Table~\ref{tab:app_rq3_splits}) so that researchers with comparable database access can reconstruct an equivalent training pool. Because all reported evaluation is conducted on PDP-Bench, the benchmark portion of our results remains independently reproducible from publicly available resources.

\begin{table}[!htbp]
\centering
\small
\setlength{\tabcolsep}{4pt}
\resizebox{\columnwidth}{!}{%
\begin{tabular}{@{}lcccc@{}}
\toprule
\textbf{Training Group} & \textbf{IENP} & \textbf{SNP} & \textbf{DNP} & \textbf{P} \\
\midrule
Balanced & 500 & 500 & 500 & 500 \\
\textsc{IENP}-40 & 800 & 400 & 400 & 400 \\
\textsc{IENP}-55 & 1{,}100 & 300 & 300 & 300 \\
\textsc{SNP}-40 & 400 & 800 & 400 & 400 \\
\textsc{SNP}-55 & 300 & 1{,}100 & 300 & 300 \\
\textsc{DNP}-40 & 400 & 400 & 800 & 400 \\
\textsc{DNP}-55 & 300 & 300 & 1{,}100 & 300 \\
\bottomrule
\end{tabular}%
}
\caption{Class counts in the \emph{pdp2k\_rq3} training splits.}
\label{tab:app_rq3_splits}
\end{table}

\subsection{DAPO Training Configuration}

All seven training groups in Section~\ref{sec:rq3} share the same training configuration; only the training split differs. Table~\ref{tab:app_dapo_config} lists the hyperparameters.

\begin{table}[!htbp]
\centering
\scriptsize
\setlength{\tabcolsep}{4pt}
\resizebox{\columnwidth}{!}{%
\begin{tabular}{@{}lll@{}}
\toprule
\textbf{Category} & \textbf{Item} & \textbf{Value} \\
\midrule
\multirow{6}{*}{Basic} & Base model & \texttt{Qwen3-8B} \\
 & RL algorithm & DAPO (TRL) \\
 & Training framework & DeepSpeed + vLLM \\
 & Numerical precision & bf16 \\
 & GPU & 2 x A100 80GB \\
 & Random seed & 42 \\
\midrule
\multirow{4}{*}{DeepSpeed} & ZeRO stage & 2 \\
 & Optimizer offload & CPU offload \\
 & Overlap communication & false \\
 & Reduce bucket size & 2e8 \\
\midrule
\multirow{4}{*}{Data} & Training split source & \emph{pdp2k\_rq3} \\
 & Evaluation split & PDP-Bench \\
 & Prompt variant & original \\
 & Training samples per split & 2,000 \\
\midrule
\multirow{9}{*}{Training} & Epochs & 4 \\
 & Per-device batch size & 8 \\
 & Gradient accumulation steps & 64 \\
 & Global effective batch size & 8 $\times$ 2 $\times$ 64 = 1,024 \\
 & Learning rate & $10^{-5}$ \\
 & Max prompt length & 1,536 \\
 & Max completion length & 1,536 \\
 & Logging steps & 2 \\
 & Save steps & 30 \\
\midrule
\multirow{5}{*}{DAPO} & Rollout generations per prompt & 8 \\
 & $\epsilon$ (low clip) & 0.2 \\
 & $\epsilon_{\text{high}}$ (high clip) & 0.28 \\
 & vLLM GPU memory utilization & 0.50 \\
 & vLLM tensor parallel size & 2 \\
\midrule
\multirow{3}{*}{Reward} & Valid format and correct decision & 1.0 \\
 & Valid format but wrong decision & 0.1 \\
 & Invalid format or unparsable decision & 0.0 \\
\bottomrule
\end{tabular}%
}
\caption{DAPO training configuration used in Section~\ref{sec:rq3}.}
\label{tab:app_dapo_config}
\end{table}

During DAPO training, the 8 rollouts for each prompt are used to estimate group-relative advantages for policy-gradient updates. Following the original DAPO setting, $\epsilon_{\text{high}}$ is larger than $\epsilon$ to reduce PPO clipping on high-advantage exploration directions.

\section{Full Class-Augmented RLVR Results}
\label{sec:app_rq3_full}

Tables~\ref{tab:app_rq3_l1} and~\ref{tab:app_rq3_l2} report the full overall and class-level F1-score results for the un-tuned Qwen3-8B baseline (\emph{no train}) and the seven \emph{pdp2k\_rq3} training splits (Balanced 25\%, and IENP/SNP/DNP at 40\% and 55\%) in Section~\ref{sec:rq3}. Values are means over 8 runs; the number after $\pm$ is the standard error computed from run-level metrics (SE, $n=8$, $df=7$).

\begin{table*}[t]
\centering
\scriptsize
\resizebox{\textwidth}{!}{%
\begin{tabular}{@{}lcccc@{}}
\toprule
\textbf{Setting} & \textbf{Macro-F1} & \textbf{Micro-F1} & \textbf{NP} & \textbf{P} \\
\midrule
no train & 0.6160 $\pm$ 0.0047 & 0.8657 $\pm$ 0.0010 & 0.3065 $\pm$ 0.0091 & 0.9255 $\pm$ 0.0005 \\
Balanced 25\% & 0.6438 $\pm$ 0.0026 & 0.8256 $\pm$ 0.0017 & 0.3895 $\pm$ 0.0042 & 0.8980 $\pm$ 0.0011 \\
\midrule
\textsc{IENP} @ 40\% & 0.4690 $\pm$ 0.0041 & 0.5163 $\pm$ 0.0060 & 0.3114 $\pm$ 0.0023 & 0.6267 $\pm$ 0.0069 \\
\textsc{IENP} @ 55\% & 0.5638 $\pm$ 0.0031 & 0.7037 $\pm$ 0.0043 & 0.3171 $\pm$ 0.0033 & 0.8105 $\pm$ 0.0034 \\
\textsc{SNP} @ 40\% & 0.6222 $\pm$ 0.0027 & 0.8133 $\pm$ 0.0025 & 0.3540 $\pm$ 0.0048 & 0.8905 $\pm$ 0.0018 \\
\textsc{SNP} @ 55\% & 0.6275 $\pm$ 0.0034 & 0.7948 $\pm$ 0.0020 & 0.3783 $\pm$ 0.0061 & 0.8767 $\pm$ 0.0013 \\
\textsc{DNP} @ 40\% & 0.4640 $\pm$ 0.0023 & 0.4985 $\pm$ 0.0034 & 0.3289 $\pm$ 0.0011 & 0.5990 $\pm$ 0.0042 \\
\textsc{DNP} @ 55\% & 0.3792 $\pm$ 0.0038 & 0.3909 $\pm$ 0.0045 & 0.2959 $\pm$ 0.0015 & 0.4625 $\pm$ 0.0063 \\
\bottomrule
\end{tabular}%
}
\caption{L1 results for the un-tuned baseline and the seven class-augmented RLVR splits.}
\label{tab:app_rq3_l1}
\end{table*}

\begin{table*}[t]
\centering
\scriptsize
\resizebox{\textwidth}{!}{%
\begin{tabular}{@{}lccccc@{}}
\toprule
\textbf{Setting} & \textbf{Macro-F1} & \textbf{Micro-F1} & \textbf{IENP} & \textbf{SNP} & \textbf{DNP} \\
\midrule
no train & 0.4546 $\pm$ 0.0051 & 0.8601 $\pm$ 0.0009 & 0.3440 $\pm$ 0.0126 & 0.3273 $\pm$ 0.0133 & 0.2216 $\pm$ 0.0114 \\
Balanced 25\% & 0.4584 $\pm$ 0.0046 & 0.8135 $\pm$ 0.0019 & 0.4467 $\pm$ 0.0149 & 0.1972 $\pm$ 0.0090 & 0.2917 $\pm$ 0.0050 \\
\midrule
\textsc{IENP} @ 40\% & 0.2745 $\pm$ 0.0017 & 0.4656 $\pm$ 0.0064 & \textbf{0.2226 $\pm$ 0.0046} & 0.0292 $\pm$ 0.0022 & 0.2194 $\pm$ 0.0035 \\
\textsc{IENP} @ 55\% & 0.3009 $\pm$ 0.0040 & 0.6660 $\pm$ 0.0046 & \textbf{0.1432 $\pm$ 0.0026} & 0.1016 $\pm$ 0.0127 & 0.1482 $\pm$ 0.0050 \\
\textsc{SNP} @ 40\% & 0.4474 $\pm$ 0.0038 & 0.8033 $\pm$ 0.0025 & 0.4092 $\pm$ 0.0108 & \textbf{0.2122 $\pm$ 0.0108} & 0.2778 $\pm$ 0.0063 \\
\textsc{SNP} @ 55\% & 0.4391 $\pm$ 0.0067 & 0.7799 $\pm$ 0.0023 & 0.4272 $\pm$ 0.0198 & \textbf{0.1596 $\pm$ 0.0089} & 0.2928 $\pm$ 0.0063 \\
\textsc{DNP} @ 40\% & 0.3956 $\pm$ 0.0054 & 0.4826 $\pm$ 0.0036 & 0.4822 $\pm$ 0.0105 & 0.2257 $\pm$ 0.0193 & \textbf{0.2756 $\pm$ 0.0009} \\
\textsc{DNP} @ 55\% & 0.3280 $\pm$ 0.0052 & 0.3714 $\pm$ 0.0047 & 0.4100 $\pm$ 0.0130 & 0.1948 $\pm$ 0.0120 & \textbf{0.2449 $\pm$ 0.0014} \\
\bottomrule
\end{tabular}%
}
\caption{L2 results for the un-tuned baseline and the seven class-augmented RLVR splits. Bold marks the target class within each intervention row.}
\label{tab:app_rq3_l2}
\end{table*}

The full estimates reinforce the pattern in Section~\ref{sec:rq3}: from the un-tuned baseline (L2 Macro-F1 0.4546), class-augmented RLVR fails to improve target-class F1 even when given strong prior intervention. On IENP and DNP, training inflates the target class's recall (Figure~\ref{fig:rq3_class_prior}) while degrading its F1 below the baseline; on SNP, even the 55\% intervention leaves F1 at 0.1596, about half of the no-train 0.3273.

\section{Law-Article Citation Scores}
\label{sec:app_articles}

This section reports auxiliary law-article citation scores from the result files. RQ1 is evaluated on PDP-Bench with a single inference run and therefore reports precision, recall, and F1. RQ2.1--RQ2.3 use 8 repeated runs; we report law-article citation F1 with standard error.

\subsection{RQ1: SOTA Models on PDP-Bench}

\begin{table}[!htbp]
\centering
\footnotesize
\setlength{\tabcolsep}{4pt}
\resizebox{\columnwidth}{!}{%
\begin{tabular}{@{}lccc@{}}
\toprule
\textbf{Model} & \textbf{Precision} & \textbf{Recall} & \textbf{F1} \\
\midrule
Claude-Opus-4.6 & 0.2667 & 0.4874 & 0.3361 \\
DeepSeek-V4-Pro & 0.3162 & 0.3669 & 0.3297 \\
Gemini-3.1-Pro & 0.3867 & 0.4954 & 0.4240 \\
GPT-5.4 & 0.2886 & 0.5151 & 0.3614 \\
GPT-OSS-20B & 0.0036 & 0.0046 & 0.0038 \\
Qwen3.5-35B-A3B & 0.1599 & 0.3623 & 0.2124 \\
Qwen3.6-Max & 0.3307 & 0.5284 & 0.3982 \\
\bottomrule
\end{tabular}%
}
\caption{Law-article citation scores for available RQ1 PDP runs.}
\label{tab:app_articles_rq1}
\end{table}

\subsection{RQ2.1: Test-Time Scaling}

\begin{table}[!htbp]
\centering
\scriptsize
\setlength{\tabcolsep}{4pt}
\resizebox{\columnwidth}{!}{%
\begin{tabular}{@{}llc@{}}
\toprule
\textbf{Model} & \textbf{Budget} & \textbf{Articles F1} \\
\midrule
\multirow{5}{*}{GPT-5.4} & none & 0.2120 $\pm$ 0.0087 \\
 & low & 0.2197 $\pm$ 0.0108 \\
 & medium & 0.2237 $\pm$ 0.0158 \\
 & high & 0.2152 $\pm$ 0.0148 \\
 & xhigh & 0.1319 $\pm$ 0.0151 \\
\midrule
\multirow{3}{*}{Gemini-3.1-Pro} & low & 0.3000 $\pm$ 0.0260 \\
 & medium & 0.2824 $\pm$ 0.0270 \\
 & high & 0.2693 $\pm$ 0.0250 \\
\midrule
\multirow{4}{*}{Claude-Opus-4.6} & 0 & 0.2067 $\pm$ 0.0032 \\
 & 1024 & 0.2824 $\pm$ 0.0034 \\
 & 2048 & 0.2776 $\pm$ 0.0038 \\
 & 4096 & 0.2746 $\pm$ 0.0039 \\
\midrule
\multirow{3}{*}{DeepSeek-V4-Pro} & none & 0.2351 $\pm$ 0.0055 \\
 & high & 0.2353 $\pm$ 0.0064 \\
 & xhigh & 0.2400 $\pm$ 0.0077 \\
\midrule
\multirow{5}{*}{Qwen3.6-Max} & 0 & 0.1946 $\pm$ 0.0104 \\
 & 512 & 0.2284 $\pm$ 0.0052 \\
 & 1024 & 0.2212 $\pm$ 0.0056 \\
 & 2048 & 0.2224 $\pm$ 0.0051 \\
 & 4096 & 0.2777 $\pm$ 0.0040 \\
\bottomrule
\end{tabular}%
}
\caption{Law-article citation F1 under different reasoning budgets.}
\label{tab:app_articles_rq21}
\end{table}

\subsection{RQ2.2: Legal-Domain Specialization}

\begin{table}[!htbp]
\centering
\footnotesize
\setlength{\tabcolsep}{4pt}
\resizebox{\columnwidth}{!}{%
\begin{tabular}{@{}llc@{}}
\toprule
\textbf{Training Paradigm} & \textbf{Model} & \textbf{Articles F1} \\
\midrule
\multirow{2}{*}{Full SFT} & Qwen2.5-7B-Instruct & 0.0085 $\pm$ 0.0055 \\
 & LawLLM-7B & 0.0209 $\pm$ 0.0109 \\
\midrule
\multirow{2}{*}{R1-Distill SFT+GRPO} & Qwen2.5-14B-Instruct & 0.0744 $\pm$ 0.0230 \\
 & LegalDelta & 0.0002 $\pm$ 0.0002 \\
\midrule
\multirow{3}{*}{Mid-training+LEAD-SFT+RL} & Qwen3-8B-Base & 0.1423 $\pm$ 0.0317 \\
 & Qwen3-8B & 0.0322 $\pm$ 0.0015 \\
 & LegalOne-R1-8B & 0.0834 $\pm$ 0.0039 \\
\bottomrule
\end{tabular}%
}
\caption{Law-article citation F1 for legal-domain specialization experiments.}
\label{tab:app_articles_rq22}
\end{table}

\subsection{RQ2.3: Prompt Augmentation}

\begin{table}[!htbp]
\centering
\footnotesize
\setlength{\tabcolsep}{3pt}
\resizebox{\columnwidth}{!}{%
\begin{tabular}{@{}lccc@{}}
\toprule
\textbf{Model} & \textbf{Baseline} & \textbf{+Definitions} & \textbf{+One-shot} \\
\midrule
GPT-5.4 & 0.2331 $\pm$ 0.0026 & 0.2420 $\pm$ 0.0038 & 0.2153 $\pm$ 0.0049 \\
Gemini-3.1-Pro & 0.3245 $\pm$ 0.0045 & 0.3632 $\pm$ 0.0070 & 0.4179 $\pm$ 0.0026 \\
Claude-Opus-4.6 & 0.2127 $\pm$ 0.0018 & 0.2738 $\pm$ 0.0022 & 0.2922 $\pm$ 0.0032 \\
DeepSeek-V4-Pro & 0.2263 $\pm$ 0.0071 & 0.2576 $\pm$ 0.0057 & 0.3647 $\pm$ 0.0056 \\
Qwen3.6-Max & 0.2627 $\pm$ 0.0055 & 0.3402 $\pm$ 0.0048 & 0.3719 $\pm$ 0.0062 \\
\bottomrule
\end{tabular}%
}
\caption{Law-article citation F1 under prompt-augmentation conditions.}
\label{tab:app_articles_rq23}
\end{table}

Overall, law-article citation remains a secondary and only partially aligned signal. Prompt augmentation improves citation F1 for most models, especially under +One-shot, but these gains do not always coincide with the best decision-prediction scores. This suggests that better citation behavior does not by itself resolve the harder liability-boundary distinctions in PDP.

\section{License of PDP-Bench}
Use of PDP-Bench is subject to the Research-only License.
